\documentclass[runningheads]{llncs}

\usepackage{eccv}
\usepackage{wrapfig}
\usepackage[table]{xcolor}

\usepackage{eccvabbrv}

\usepackage{graphicx}
\usepackage{booktabs}

\usepackage[accsupp]{axessibility}  

\usepackage{hyperref}

\usepackage{orcidlink}

\begin{document}

\title{BridgeDiff: Bridging Human Observations and Flat-Garment Synthesis for Virtual Try-Off} 

\titlerunning{BridgeDiff for Virtual Try-Off}

\author{
Shuang Liu\inst{1} \and
Ao Yu\inst{1} \and
Linkang Cheng\inst{1} \and
Xiwen Huang\inst{1} \and
Li Zhao\inst{1} \and
Junhui Liu\inst{3} \and
Zhiting Lin\inst{1,2} \and
Yu Liu\inst{1,2}\thanks{Corresponding authors}
}

\authorrunning{S. Liu et al.}

\institute{
School of Integrated Circuits, Anhui University, Hefei, China
\and
Anhui Provincial High-performance Integrated Circuit Engineering Research Center
\and
School of Astronautics, Northwestern Polytechnical University, Xi'an,China
}

\maketitle

\begin{abstract}
  Virtual try-off (VTOFF) aims to recover canonical flat-garment representations from images of dressed persons for standardized display and downstream virtual try-on. Prior methods often treat VTOFF as direct image translation driven by local masks or text-only prompts, overlooking the gap between on-body appearances and flat layouts. This gap frequently leads to inconsistent completion in unobserved regions and unstable garment structure. We propose BridgeDiff, a diffusion-based framework that explicitly bridges human-centric observations and flat-garment synthesis through two complementary components. First, the Garment Condition Bridge Module (GCBM) builds a garment-cue representation that captures global appearance and semantic identity, enabling robust inference of continuous details under partial visibility. Second, the Flat Structure Constraint Module (FSCM) injects explicit flat-garment structural priors via Flat-Constraint Attention (FC-Attention) at selected denoising stages, improving structural stability beyond text-only conditioning. Extensive experiments on standard VTOFF benchmarks show that BridgeDiff achieves state-of-the-art performance, producing higher-quality flat-garment reconstructions while preserving fine-grained appearance and structural integrity.
\end{abstract}

\section{Introduction}
\label{sec:intro}

In the rapidly evolving landscape of e-commerce, particularly within the fashion industry, providing realistic, flexible, and scalable visual representations of garments has become increasingly important for enhancing user experience and driving consumer engagement. While virtual try-on (VTON) \cite{choi2021viton,choi2024improving,guo2025any2anytryon,lee2025voost,lee2022high,shen2025imagdressing,chong2024catvton,kim2024stableviton} techniques enable customers to visualize how a garment appears when worn by a person, they typically rely on standardized catalog images that are costly to produce. To address this limitation, the recently proposed virtual try-off (VTOFF) \cite{velioglu2024tryoffdiff,velioglu2025mgt,xarchakos2024tryoffanyone,guo2025any2anytryon,lee2025voost,liu2025one,zhang2025unifit} task aims to reconstruct clean, canonical flat-garment representations directly from images of dressed individuals. By recovering garments in a standardized, catalog-style format, VTOFF offers significant value for product presentation, visual retrieval, and downstream applications (e.g., person-to-person try-on). However, accurately reconstructing flat-garments from on-body observations remains challenging due to severe occlusions, pose variations, and the inherent distribution gap between human-centric images and canonical flat-garment layouts. 
Effectively bridging this gap while ensuring visual continuity of garment details and maintaining stable garment structure under limited or partially observable conditions is therefore a key factor for advancing VTOFF performance.

Existing studies on VTOFF can be broadly categorized into two lines of research. The first line focuses on standalone VTOFF modeling, formulating the task as reconstructing clean and canonical flat-garment images directly from photos of dressed person. Representative works include TryOffDiff \cite{velioglu2024tryoffdiff}, which pioneers diffusion-based garment reconstruction, MGT \cite{velioglu2025mgt}, which introduces category-specific embeddings to support multiple garment types, and TryOffAnyone \cite{xarchakos2024tryoffanyone}, which leverages explicit garment masks and lightweight fine-tuning to balance efficiency and quality. The second line of research aims to unify VTON and VTOFF within a single framework to improve generality and scalability. Methods including Any2AnyTryOn \cite{guo2025any2anytryon} and Voost \cite{lee2025voost} adopt Diffusion Transformer architectures with task-aware conditioning, while OMFA \cite{liu2025one} and UniFit \cite{zhang2025unifit} further enhance human–garment disentanglement and semantic alignment under diverse poses and multimodal instructions. Despite these advances, most existing VTOFF methods still formulate the task as a direct mapping from dressed-person images to flat-garment images, typically relying on local mask constraints or coarse textual cues. Such designs struggle to handle partially observable regions, often leading to discontinuous or implausible garment details under occlusions and pose variations.
Moreover, although unified multi-task frameworks improve flexibility, their reliance on text-dominant conditioning makes it difficult to consistently enforce stable flat-garment structures during generation. These limitations highlight the need for explicitly modeling garment cues representation and introducing structured constraints to ensure both visual continuity and structural stability in VTOFF.
\begin{figure*}[t]
    \centering
    \includegraphics[width=\textwidth,
                 trim=0.5cm 0.6cm 0.5cm 0.8cm,
                 clip]{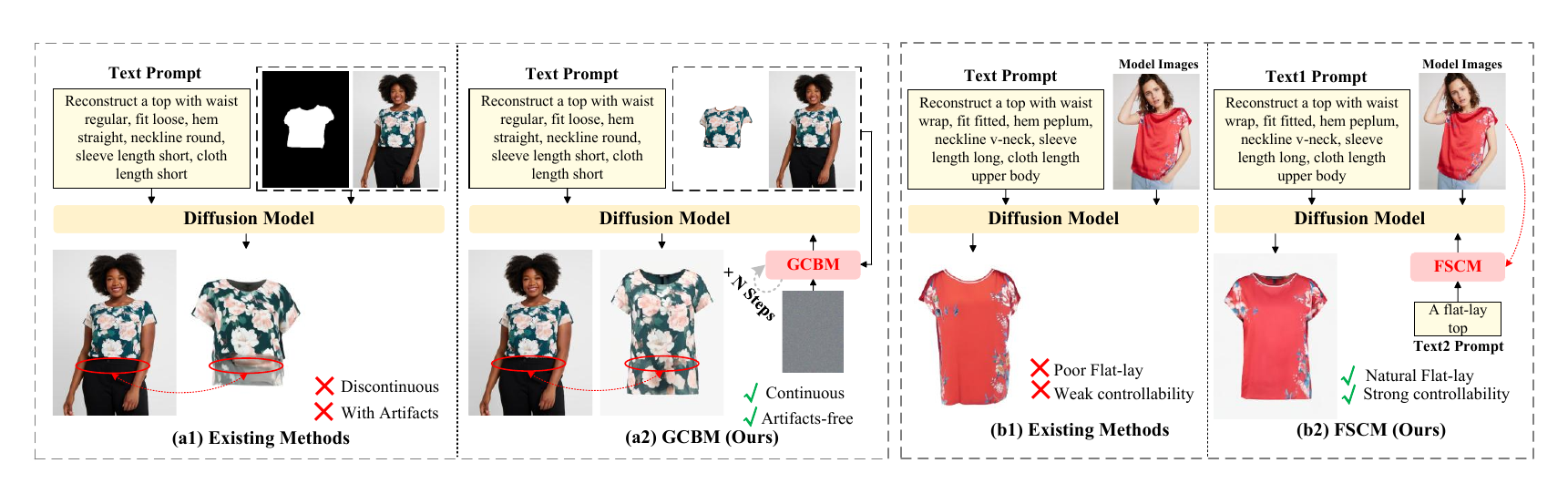}
    \caption{
    Existing methods often suffer from visual discontinuity and structural instability when relying solely on textual conditioning or weak local constraints, especially under occlusions or partial observations.
    In contrast, BridgeDiff bridges dressed-person observations and canonical flat-garment representations via garment cues representation and explicit flat structure guidance.
    }
    \label{fig:motivation}
\end{figure*}

Motivated by these observations, we propose BridgeDiff, a structure-aware diffusion framework for the VTOFF task, aiming to generate visually continuous and structurally stable flat-garments from human-centric observation images. As illustrated in Fig.~\ref{fig:motivation}, existing VTOFF methods often suffer from two key limitations.
First, under occlusions or partial observations, the absence of garment cues representation makes it difficult to infer visually continuous details for unseen regions.
Second, relying solely on textual descriptions or weak local constraints is insufficient to enforce stable flat-garment geometry during generation. To address these challenges, BridgeDiff explicitly bridges the gap between dressed-person observations and canonical flat garment representations by introducing garment cues representation and explicit flat structure guidance within the diffusion process. Specifically, we design a garment condition bridge module (GCBM) to aggregate global garment appearance from dressed-person images into garment cues representation, facilitating continuous inference of garment details beyond visible regions.
In addition, we introduce a flat structure constraint module (FSCM), which injects explicit flat-garment structural information into a specific stage of the diffusion denoising UNet, further improving the structural stability of the generated results.
\noindent\textbf{Our contributions are summarized as follows:}
\begin{itemize}
    \item We propose a garment condition bridge module (GCBM) that constructs garment cues representation, enabling stable modeling of global garment appearance and semantic identity, and supporting visually continuous detail generation under limited conditioning in VTOFF scenarios.
    \item We introduce flat structure constraint module (FSCM) that injects explicit flat-garment structural information into the diffusion denoising process via structure-aware attention, significantly improving structural stability and geometric plausibility.
    \item We conduct comprehensive experiments on two public VTOFF benchmarks, together with a user study, demonstrating the effectiveness of BridgeDiff in terms of both quantitative performance and visual quality.
\end{itemize}

\section{Related Work}\label{sec:rw}

\noindent\textbf{Virtual Try-On.}
Image-based virtual try-on (VTON) synthesizes a person wearing a target garment while preserving pose, body shape, and identity.
Early VTON methods are largely GAN-based~\cite{wang2018toward,choi2021viton,lee2022high,goodfellow2020generative}, which often suffer from training instability and limited fidelity on fine garment details.
Diffusion models~\cite{sohl2015deep,ho2020denoising,shen2024imagpose,shen2024advancing} have recently become the dominant paradigm for VTON~\cite{morelli2023ladi,gou2023taming,chong2024catvton,shen2025imagdressing,choi2024improving,yang2025omnivton} due to their strong generation quality.
Early diffusion-based approaches commonly follow warping-based pipelines~\cite{morelli2023ladi,gou2023taming}, which can introduce artifacts under imperfect geometric alignment.
Warping-free diffusion methods~\cite{zhu2023tryondiffusion,xu2025ootdiffusion,choi2024improving} and designs with garment encoders or dual UNet architectures~\cite{kim2024stableviton,shen2025imaggarment,shen2025imagdressing} further improve clothing fidelity.
However, most VTON work focuses on the forward synthesis setting, while the inverse setting remains less studied.

\noindent\textbf{Virtual Try-Off.}
It aims to reconstruct clean garment representations from images of dressed persons.
TryOffDiff~\cite{velioglu2024tryoffdiff} initiates this direction with diffusion-based garment generation, and MGT~\cite{velioglu2025mgt} extends it with category-aware embeddings for multi-garment modeling.
TryOffAnyone~\cite{xarchakos2024tryoffanyone} incorporates garment masks and parameter-efficient tuning to balance quality and efficiency.
Recent works further move toward unifying VTON and VTOFF.
Any2AnyTryOn~\cite{guo2025any2anytryon} and Voost~\cite{lee2025voost} adopt DiT backbones~\cite{peebles2023scalable} with task conditioning for scalable multi-task learning, while OMFA~\cite{liu2025one} and UniFit~\cite{zhang2025unifit} explore bidirectional generation and semantic alignment via richer conditioning.
Despite this progress, most VTOFF methods regress flat garments directly from dressed-person images using category text or local masks, while neglecting the distribution gap between occluded, pose-dependent observations and canonical flat-garment structure. This often yields discontinuous details and unstable geometry.

\section{Methodology}
\subsection{Garment Condition Bridge Module} 

A fundamental challenge in VTOFF lies in the distribution gap between human-centric observation images and canonical flat-garment synthesis space. In conventional VTON tasks, this gap can be largely mitigated by leveraging rich human-related conditions (e.g., pose information, human parsing maps, and garment masks), which help models focus on the target garment regions. 
\begin{wrapfigure}{r}{0.46\linewidth}
    \centering
    \includegraphics[width=\linewidth,
                     trim=0.6cm 0.6cm 0.8cm 0.7cm,
                     clip]{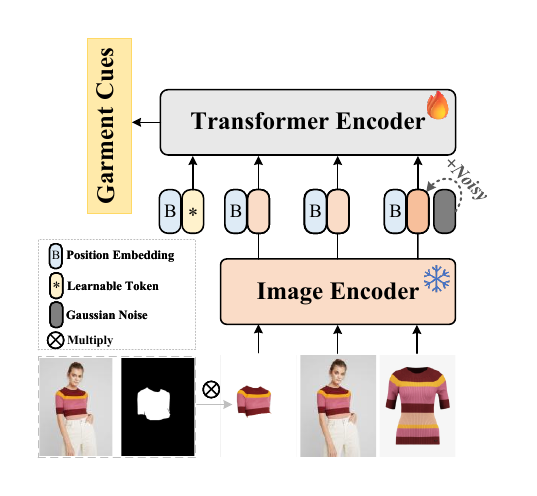}
    \caption{Overview of the proposed GCBM.
Rather than directly mapping dressed-person observations to flat garment images, GCBM aggregates multiple visual information into garment cues representation, capturing the global appearance and identity of the target garment to support visually continuous flat-garment synthesis.}
    \label{fig:gcbm}
\end{wrapfigure}
However, in VTOFF scenarios, many of these conditions are unavailable or unreliable, making it difficult for the model to accurately recover visually continuous details in occluded regions under sparse or partial constraints. 
To address this challenge, we propose explicitly constructing garment cues representation that captures the global appearance and semantic identity of the target garment.

Motivated by the hierarchical representation modeling strategy adopted in DALL·E 2~\cite{ramesh2022hierarchical}, we introduce the GCBM to construct garment cues representation for VTOFF. As illustrated in Fig.~\ref{fig:gcbm}, we first apply a cloth-agnostic mask to the reference model image $X_m$ to obtain a warped garment representation $X_w$.
Then, the warped garment image $X_w$, the reference model image $X_m$, and the corresponding target garment image $X_c$ are independently encoded by a shared image encoder, producing feature representations $F_w$, $F_m$, and $F_c$, respectively.
To align with the diffusion timestep, the target garment feature $F_c$ is perturbed through a forward noising process \footnote{In our experiments, we adopt the forward noising process defined in DDPM~\cite{ho2020denoising}. The implementation is based on \url{https://github.com/hojonathanho/diffusion}.}, yielding a noisy feature representation denoted as $\hat{F}_c$. Subsequently, The feature tokens $F_w$, $F_m$, and $\hat{F}_c$ are concatenated along the token dimension and fed into the GCBM.
The internal architecture of GCBM follows MetaFormer\cite{yu2022metaformer} design, enabling flexible token mixing across heterogeneous feature sources.
To distinguish different conditional features, we use learnable positional encodings for each type, enabling effective differentiation of different conditional feature during aggregation.
The training objective of the GCBM, parameterized by $\theta$, is defined as follows:
\begin{equation}
L_{\text{GCBM}} =
\mathbb{E}_{F_c, F_w, F_m, \epsilon, t}
\Big[
\big\| F_c - f_\theta(\hat{F}_c, F_w, F_m, t) \big\|_2^2
\Big],
\end{equation}
where $t$ denotes the diffusion timestep. At inference time, GCBM starts from a pure noise initialization and iteratively denoises it to progressively refine the target garment cues representation. At each diffusion step $t$, the model conditions on the model image feature $F_m$ and the warped garment representation $F_w$, and predicts the clean garment cues representation as:
\begin{equation}
F_c = f_\theta(\epsilon, F_w, F_m, t),
\end{equation}
where $\epsilon$ denotes the gaussian noise.
After completing the iterative denoising process \footnote{At inference time, we adopt the DDIM \cite{song2020denoising} reverse process for efficient sampling, following the implementation in \url{https://github.com/ermongroup/ddim}.}, the model obtains approximately garment cues.
This cues serves as a semantic bridge between human-centric observations and the canonical flat garment layout, and is subsequently used as a conditioning signal for downstream diffusion-based synthesis.

\subsection{Flat Structure Constraint for Conditional Diffusion} 
\begin{figure}[tb]
  \centering
  \includegraphics[width=\textwidth,
                 trim=0.8cm 0.6cm 0.7cm 0.8cm,
                 clip]{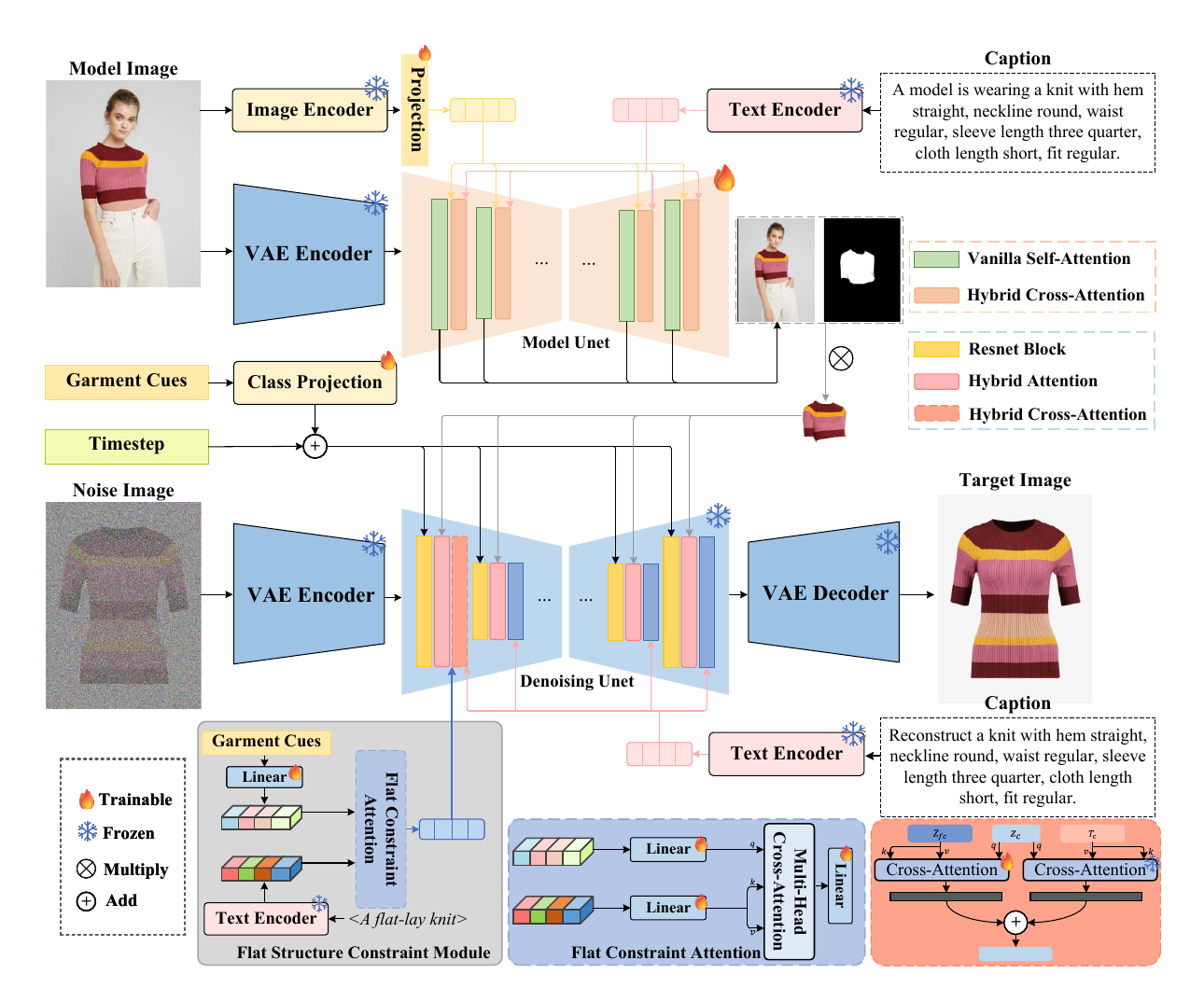}
  \caption{Overview of the proposed Flat Structure Constraint for Conditional Diffusion architecture. The framework consists of a trainable model UNet and a largely frozen denoising UNet. 
To explicitly enforce flat garment layouts, a flat structure constraint module (FSCM) is integrated into the denoising UNet, ensuring stable layout generation without compromising appearance fidelity.
}
  \label{fig:unet}
\end{figure}

\noindent\textbf{Overview.} Our goal is to generate structurally stable flat garment images while preserving fine-grained appearance details. To this end, we propose a Flat Structure Constraint for Conditional Diffusion framework to explicitly enforce the flat layout of garments.
As illustrated in Fig.~\ref{fig:unet}, the proposed framework consists of three main components: a model UNet, a denoising UNet, and a FSCM. The parameters of the denoising UNet are largely frozen, and  FSCM is incorporated into it to maintain the structural stability of the generated flat garments. Meanwhile, the model UNet serves as a conditional feature extractor, which captures fine-grained garment details from the input model image.

\noindent\textbf{Model Unet.} Unlike conventional diffusion frameworks that rely solely on image encoders for conditional guidance, the model UNet directly processes the input model image and its textual description to extract intermediate garment features. During both training and inference, the latent variables are kept noise-free and a single forward pass is performed at diffusion timestep $T=0$, following the design adopted in IMAGDressing-v1 \cite{shen2025imagdressing} and IDM-VTON \cite{choi2024improving}. The extracted intermediate features are then used as conditional guidance for the denoising UNet.
Specifically, given a model image $X_m \in \mathbb{R}^{3 \times H \times W}$, a frozen Variational Autoencoder (VAE) encoder and an image encoder are applied to obtain the latent representation
$Z_m \in \mathbb{R}^{4 \times \frac{H}{8} \times \frac{W}{8}}$ and the image embedding
$I_m \in \mathbb{R}^{1024}$, respectively. The corresponding textual description is encoded by a text encoder to produce $T_m \in \mathbb{R}^{77 \times 768}$. The image embedding $I_m \in \mathbb{R}^{1024}$ is projected to $I'_m \in \mathbb{R}^{768}$, and the hybrid cross-attention at the $i$-th layer is defined as:
\begin{equation}
Z_{\mathrm{hc}}^{(i)} =
\mathrm{Softmax}\!\left(\frac{Q^{(i)}K_T^\top}{\sqrt{d}}\right) V_T
+ \lambda \,
\mathrm{Softmax}\!\left(\frac{Q^{(i)}{K_I}^\top}{\sqrt{d}}\right) V_I,
\end{equation}
where $\lambda$ is a hyperparameter used to regulate the contribution of image-conditioned attention.
Specifically, the query, key, and value matrices are defined as
$Q^{(i)} = Z_m^{(i)} W_q$,
$K_t = T_m W_k$,
and $V_t = T_m W_v$ for the text branch,
while $K_e = I'_m W_k$ and $V_e = I'_m W_v'$ correspond to the image branch.
The projection matrices $W_k'$ and $W_v'$ are initialized from $W_k$ and $W_v$, respectively, and remain learnable during training.
\begin{table}[t]
\centering
\caption{Quantitative results on the DressCode dataset over the full test set and category-specific subsets.
$\uparrow$ indicates higher is better, $\downarrow$ indicates lower is better. 
\textbf{Bold} denotes the best result and \underline{underline} denotes the second-best result. $*$ denotes results obtained using a single unified model trained jointly on all three garment categories.}

\label{tab:dresscode_quant}
\resizebox{\textwidth}{!}{
\begin{tabular}{lcccccc c cccccc}
\toprule
 & \multicolumn{6}{c}{All} & & \multicolumn{6}{c}{Upper-Body} \\
\cmidrule(lr){2-7} \cmidrule(lr){9-14}
Method
& FID$\downarrow$ & KID$\downarrow$ & PSNR$\uparrow$ & SSIM$\uparrow$ & LPIPS$\downarrow$ & DISTS$\downarrow$
& 
& FID$\downarrow$ & KID$\downarrow$ & PSNR$\uparrow$ & SSIM$\uparrow$ & LPIPS$\downarrow$ & DISTS$\downarrow$ \\
\midrule
Any2AnyTryOn \cite{guo2025any2anytryon}
& 17.25 & \underline{7.46} & \underline{12.80} & 74.43 & 36.68 & \underline{24.12}
& 
& \textbf{15.29} & \textbf{2.14} & \underline{12.97} & 74.48 & \underline{37.74} & \underline{23.38} \\

MGT$^*$\cite{velioglu2025mgt}          
& \underline{12.62} & 7.56 & 12.72 & \underline{75.35} & \underline{35.01} & 24.73
& 
& 19.18 & 6.65 & 12.27 & \underline{74.38} & 40.24 & 25.42 \\

\rowcolor{blue!5}
BridgeDiff (Ours)   
& \textbf{10.92} & \textbf{3.86} & \textbf{16.69} & \textbf{80.23} & \textbf{23.10} & \textbf{20.82}
& 
& \underline{17.60} & \underline{4.96} & \textbf{15.90} & \textbf{79.07} & \textbf{27.52} & \textbf{22.36} \\
\midrule

 & \multicolumn{6}{c}{Lower-Body} & & \multicolumn{6}{c}{Dresses} \\
\cmidrule(lr){2-7} \cmidrule(lr){9-14}
Method
& FID$\downarrow$ & KID$\downarrow$ & PSNR$\uparrow$ & SSIM$\uparrow$ & LPIPS$\downarrow$ & DISTS$\downarrow$
& 
& FID$\downarrow$ & KID$\downarrow$ & PSNR$\uparrow$ & SSIM$\uparrow$ & LPIPS$\downarrow$ & DISTS$\downarrow$ \\
\midrule
Any2AnyTryOn 
& 49.04 & 22.37 & 12.51 & 74.47 & 37.55 & 25.51
& 
& 28.78 & 14.14 & 12.91 & 74.35 & 34.74 & \underline{23.47} \\

MGT$^*$          
& \underline{22.15} & \underline{6.48} & \underline{12.54} & \underline{75.21} & \underline{34.58} & \underline{24.63}
& 
& \underline{20.35} & \underline{6.09} & \underline{13.41} & \underline{76.46} & \underline{30.22} & 24.13 \\

\rowcolor{blue!5}
BridgeDiff (Ours)   
& \textbf{19.86} & \textbf{5.13} & \textbf{16.71} & \textbf{80.99} & \textbf{22.71} & \textbf{21.08}
& 
& \textbf{16.37} & \textbf{4.72} & \textbf{17.52} & \textbf{80.65} & \textbf{19.07} & \textbf{19.03} \\
\bottomrule
\end{tabular}}
\vspace{-0.6cm}
\end{table}
\noindent\textbf{Denoising UNet.} In the denoising UNet, intermediate features from the model UNet, denoted as $M_o \in \mathbb{R}^{N \times C}$, are used as additional conditioning and injected into all self-attention layers of the denoising UNet. Given a cloth-agnostic mask $M_c \in \mathbb{R}^{N \times 1}$, the masked output for an intermediate layer is defined as:
\begin{equation}
M_o' = \mathrm{Interpolate}(M_c) \otimes M_o,
\end{equation}
where $\otimes$ denotes the element-wise (Hadamard) product, and $\mathrm{Interpolate}(\cdot)$ denotes the mask to match the spatial resolution of $M_o$. The resulting masked features
$M_o' \in \mathbb{R}^{N \times C}$ are then injected as additional conditioning inputs into the corresponding self-attention layers of the denoising UNet. Then, we replace all vanilla self-attention layers with hybrid attention layers to incorporate conditional features. Given a target garment image $X_c \in \mathbb{R}^{3 \times H \times W}$, we first encode it using a frozen VAE encoder to obtain the latent representation $Z_c \in \mathbb{R}^{4 \times \frac{H}{8} \times \frac{W}{8}}$. Then, Gaussian noise is added to $Z_c$ to obtain the added noisy latent $z_c$ at timestep $t$. At the $i$-th layer of the denoising UNet, the hybrid attention output $Z_{\mathrm{hs}}^{(i)}$ is defined are as follow:
\begin{equation}
Z_{\mathrm{h}}^{(i)} =
\mathrm{Softmax}\!\left(
\frac{QK^{\top}}{\sqrt{d}}
\right) V
+
\beta \,
\mathrm{Softmax}\!\left(
\frac{Q\big({K_o^{(i)}}\big)^{\top}}{\sqrt{d}}
\right) V_o^{(i)},
\end{equation}
where $\beta$ is a hyperparameter controlling the contribution of the conditioned branch. Specifically, the query, key, and value matrices of the latent self-attention branch are defined as $Q = z_c W_q$, $K = z_c W_k$ and $V = z_c W_v$ while the condition branch is constructed from the masked model UNet features $M_o'^{(i)}$ where the corresponding key and value are given by $K_o^{(i)} = M_o'^{(i)} \hat{W_k}$ and $V_o^{(i)} = M_o'^{(i)} \hat{W_v}$. The projection matrices $\hat{W_k}$ and $\hat{W_v}$ are initialized from $W_k$ and $W_v$, respectively, and remain learnable during training.

\noindent\textbf{Flat Structure Constraint Module.}\begin{figure}[t]
    \centering
    \includegraphics[width=\textwidth,
                 trim=0.5cm 0.6cm 0.7cm 0.7cm,
                 clip]{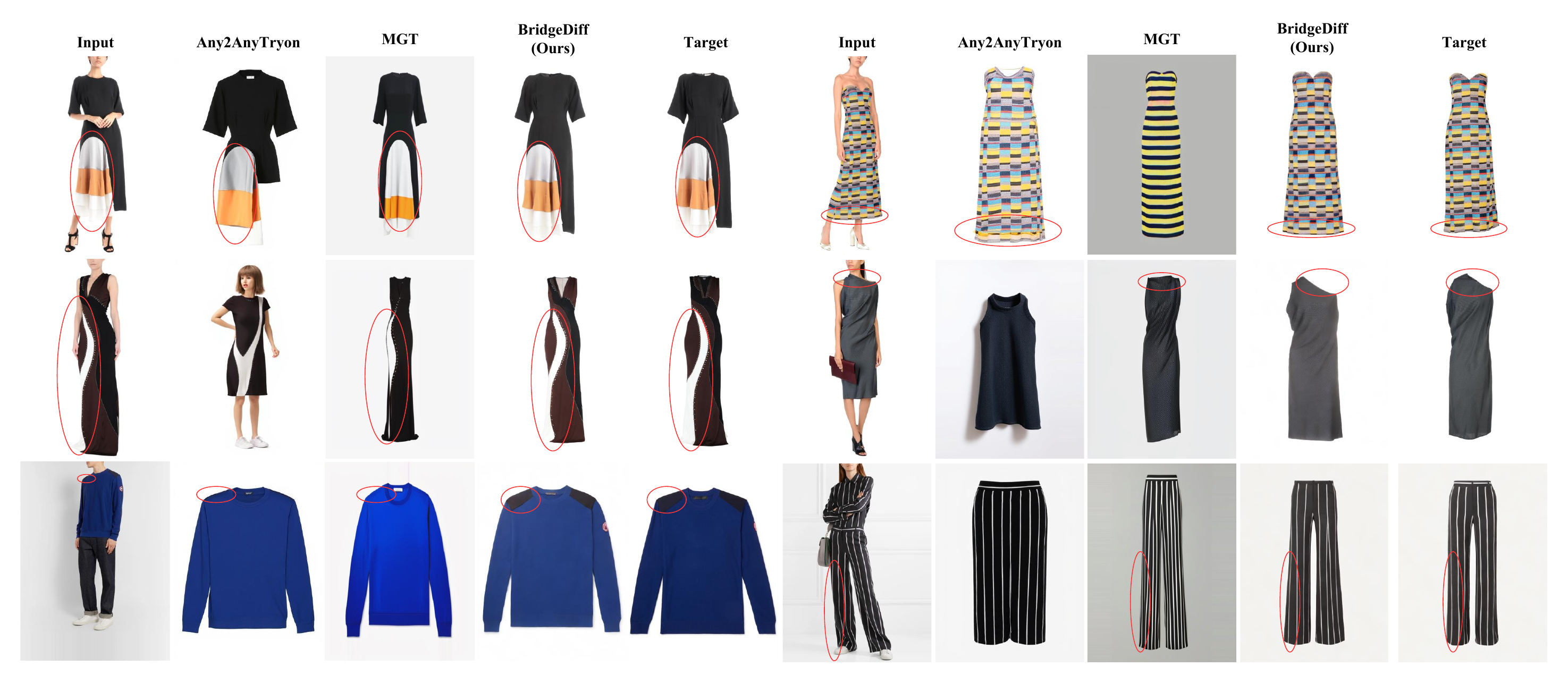}
    \caption{Qualitative comparisons on the DressCode dataset.  
\textcolor{red}{Red circles} highlight differences in local regions across different methods.  
Unmarked examples indicate cases where the overall garment structure or color appearance differs from the reference.  
\textbf{Zooming in provides a clearer view of these differences.}}
    \label{fig:dresscode_compared}
\end{figure} For flat garment generation, the canonical flat layout cannot be reliably preserved when conditioning relies solely on textual descriptions. To address this issue, we focus on decoupling semantic appearance control from structural layout constraints. We introduce the FSCM into the denoising UNet at specific stages to explicitly enforce flat garment layouts during generation. The module consists of a single flat-constraint attention (FC-Attention) layer and a learnable linear projection layer. Specifically, we extract garment clues representation from the GCBM, which are denoted as $F_c \in \mathbb{R}^{N \times 1024}$. In parallel, the corresponding flat garment textual description \footnote{Inspired by the captioning convention used in CLIP pre-training \cite{radford2021learning}, we employ a fixed textual template, ``A flat-lay \texttt{<category>},'' to describe all target flat garment images.} is processed by a text encoder, producing text embeddings $T_{flat} \in \mathbb{R}^{77 \times 768}$. To align with the dimensionality of the text embeddings, $F_c$ is further projected through a linear projection layer, resulting visual garment cues $F_c' \in \mathbb{R}^{N \times 768}$. The FC-Attention layer is designed to fuse visual garment cues with flat garment textual descriptions. For a single attention head $m$, the FC-Attention output is defined as:
\begin{equation}
z^{(m)} =
\mathrm{Softmax}\!\left(
\frac{Q^{(m)} \big(K^{(m)}\big)^\top}{\sqrt{d}} 
\right) V^{(m)},
\end{equation}
The final FC-Attention output is obtained by concatenating the outputs of all $G$ heads
and applying a learnable linear projection $W_o$:
\begin{equation}
Z = \mathrm{Concat}\big(z^{(1)}, z^{(2)}, \ldots, z^{(G)}\big) W_o,
\end{equation}
where $Q$ is projected from the visual garment cues $F_c'$, and $K$ and $V$ are projected from the flat garment textual embeddings $T_{flat}$. The resulting FC-Attention output is denoted as $Z_{fc} \in \mathbb{R}^{N \times 768}$, which subsequently injected into the denoising UNet at selected stage. Specifically, at the $j$-th cross-attention layer, $Z_{fc}$ is incorporated as an
additional conditioning signal alongside the appearance-based textual description.
Given a garment appearance prompt encoded by a text encoder as $T_c$ and the corresponding hidden feature at this layer denoted as $z_c$, The modified cross-attention output $z_j$ is defined as:
\begin{equation}
z_j =
\mathrm{Softmax}\!\left(\frac{Q K^\top}{\sqrt{d}}\right) V
\;+\;
\gamma \,
\mathrm{Softmax}\!\left(\frac{Q K_{fc}^\top}{\sqrt{d}}\right) V_{fc},
\end{equation}\begin{table}[t]
\setlength{\tabcolsep}{8pt}
\centering
\caption{Quantitative results on the VITON-HD dataset over the full test set and category-specific subsets.
$\uparrow$ indicates higher is better, $\downarrow$ indicates lower is better.
\textbf{Bold} denotes the best result and \underline{underline} denotes the second-best result.
``--'' indicates the metric is not reported.
$\dagger$ denotes cross-dataset evaluation where the model is trained on DressCode and evaluated on VITON-HD.}
\label{tab:vitonhd_quant}
\small
\begin{tabular}{lcccccc}
\toprule
Method 
& FID$\downarrow$ 
& KID$\downarrow$ 
& PSNR$\uparrow$ 
& SSIM$\uparrow$ 
& LPIPS$\downarrow$ 
& DISTS$\downarrow$ \\
\midrule
TryOffDiff \cite{velioglu2024tryoffdiff}        
& 22.54 & 8.58 & 11.77 & 71.83 & 42.27 & 24.87 \\
TryOffAnyone \cite{xarchakos2024tryoffanyone}     
& 12.59 & 2.77 & 12.51 & 71.90 & 34.47 & 22.68 \\
Any2AnyTryOn \cite{guo2025any2anytryon}     
& 12.15 & \underline{1.96} & \underline{12.86} & 74.29 & 35.72 & 22.80 \\
Voost \cite{lee2025voost}             
& \underline{10.06} & 2.48 & -     & -     & -     & -     \\
UniFit \cite{zhang2025unifit} 
& 12.58  & - & -     & \textbf{77.50} & \underline{28.10} & \underline{20.20} \\
\rowcolor{blue!5}
BridgeDiff (Ours)        
& \textbf{9.08} 
& \textbf{1.53} 
& \textbf{15.00} 
& \underline{77.42} 
& \textbf{24.38} 
& \textbf{18.69} \\
\midrule
MGT$^\dagger$ \cite{velioglu2025mgt}              
& 24.00 & 10.06 & 10.63 & 70.77 & 47.45 & 29.98 \\
\rowcolor{blue!5}
BridgeDiff (Ours)$^\dagger$        
& \textbf{17.65} & \textbf{5.75} & \textbf{12.88} & \textbf{74.24} & \textbf{35.05} & \textbf{24.79} \\
\bottomrule
\end{tabular}
\end{table}where $\gamma$ balances the contribution of structure guidance. The query is $Q = z_c W_q$.  
For the appearance-conditioned branch, $K = T_c W_k$ and $V = T_c W_v$, while for the branch conditioned on the flat structure, $K_{fc} = Z_{fc} W_k^{f}$ and $V_{fc} = Z_{fc} W_v^{f}$.  
The projection matrices $W_k^{f}$ and $W_v^{f}$ are initialized from $W_k$ and $W_v$ and remain learnable.  
All other cross-attention layers in the denoising UNet use only textual appearance features $T_c$, as in standard Text-to-Image (T2I). This injection strategy decouples semantic appearance control from structural layout constraints, preserving the native T2I generation capability of the denoising UNet. The explicit structural constraints of the FSCM strengthen flat garment structures during the diffusion process, effectively stabilizing the structures without sacrificing appearance fidelity.

\section{Experiment}
\subsection{Datasets and Metrics}
Our experiments are conducted on two public datasets: VITON-HD \cite{choi2021viton} and DressCode \cite{morelli2022dress}. VITON-HD contains 13,679 high-resolution (1024 × 768) upper-body garment image pairs, with 11,647 pairs for training and 2,032 pairs for testing. 
DressCode includes 53,792 high-resolution (1024 × 768) full-body person–garment pairs, with 48,392 pairs for training and 5,400 pairs for testing, covering upper-body, lower-body, and dress categories. To evaluate the reconstruction quality in paired try-off settings, we adopt several widely used full-reference and perceptual metrics: SSIM \cite{wang2004image}, LPIPS \cite{zhang2018unreasonable}, FID \cite{parmar2022aliased}, KID \cite{binkowski2018demystifying}, DISTS \cite{ding2020image}, and PSNR.
\begin{figure}[t]
    \centering
    \includegraphics[width=\textwidth,
                 trim=0.5cm 0.6cm 0.7cm 0.7cm,
                 clip]{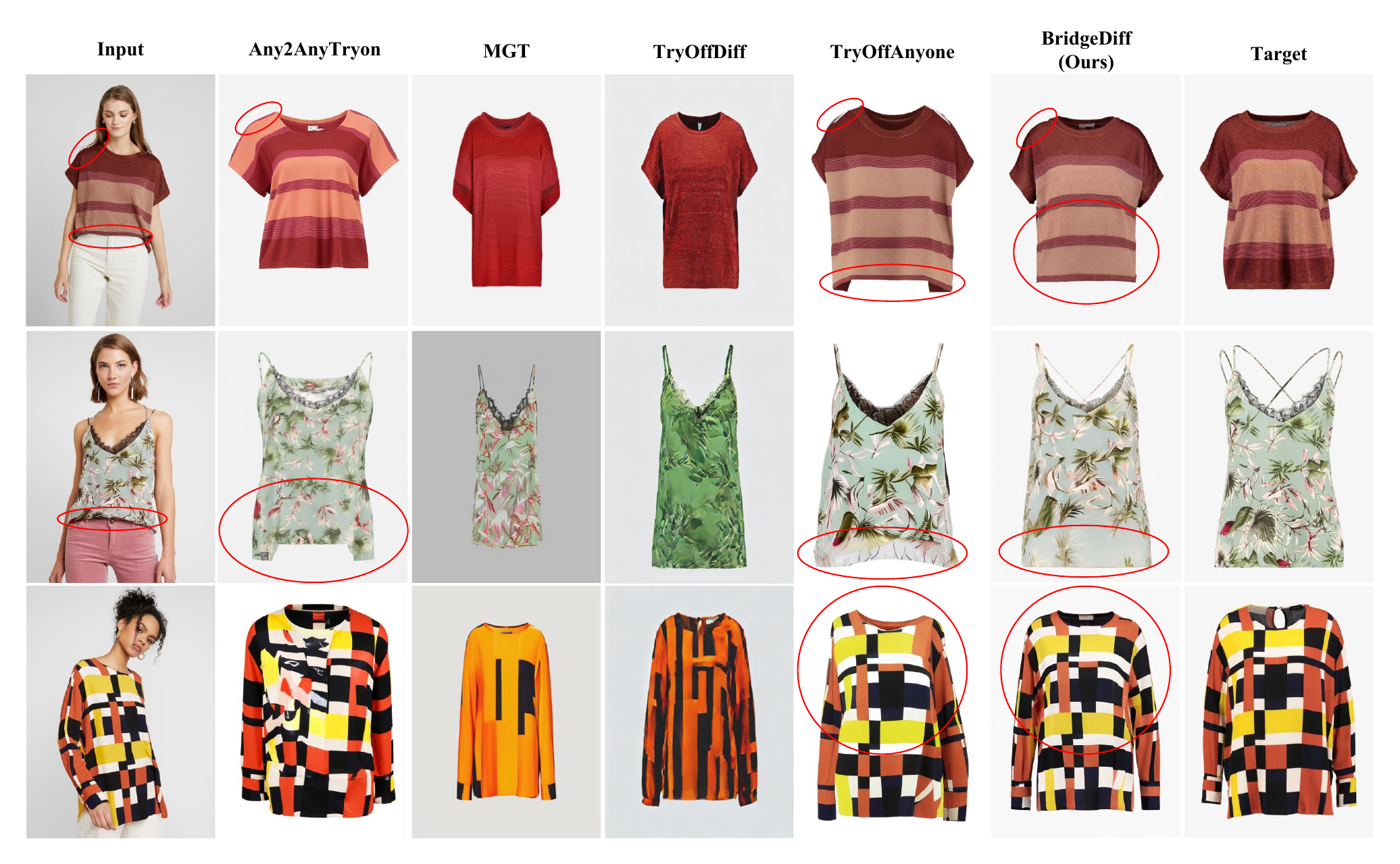}
    \caption{Qualitative comparisons on the VITON-HD dataset.
    \textcolor{red}{Red circles} highlight differences in local regions across different methods.  
Unmarked examples indicate cases where the overall garment structure or color appearance differs from the reference.  
    \textbf{Zooming in provides a clearer view of these differences.}}
    \label{fig:vitonhd_compared}
\end{figure}
\subsection{Implementation details}
Our framework is trained in a two-stage manner. In the first stage, the model follows MetaFormer \cite{yu2022metaformer} architecture. Following the design of ConvNeXt \cite{liu2022convnet}, each Feed-Forward Network (FFN) has an intermediate layer 4 times the hidden dimension. We employ OpenCLIP ViT-H/14 \cite{ilharco2021openclip} as the image encoder. In the second stage, both models are built upon the standard Stable Diffusion v1.5 \cite{rombach2022high}. For the model UNet, we employ SigLIP2 L/16-512 \cite{tschannen2025siglip} and OpenAI ViT-L/14  \cite{radford2021learning} as the image encoder and text encoder, respectively. For the denoising UNet, OpenAI ViT-L/14 is used as the text encoder. For second stage, we apply the DREAM \cite{zhou2024dream} training strategy with the parameter set to $p$ = 1. All experiments are conducted on 4 NVIDIA RTX 4090 GPUs. For each stage, we train two separate model versions on the VITON-HD and DressCode datasets independently. We use the AdamW optimizer \cite{loshchilov2017decoupled} with a constant learning rate of $1 \times 10^{-5}$ and a weight decay of $0.01$. The batch size is set per GPU, with 128 for the first stage and 8 for the second stage. Both stages are trained at a resolution of $512 \times 512$, with 65k and 54k training steps, respectively.

\subsection{Main Results} 
We compare our method with recent VTOFF approaches. VTOFF-specific methods are TryOffDiff~\cite{velioglu2024tryoffdiff}, MGT~\cite{velioglu2025mgt}, and TryOffAnyone~\cite{xarchakos2024tryoffanyone}. Unified multi-task frameworks are Voost~\cite{lee2025voost}, Any2AnyTryOn~\cite{guo2025any2anytryon}, and UniFit~\cite{zhang2025unifit}.  
Due to unavailable implementations, results for Voost and UniFit are taken from their original papers. TryOffDiff and TryOffAnyone are evaluated only on VITON-HD, while MGT is tested in a cross-dataset setting.\begin{table}[t]
\centering
\footnotesize
\setlength{\tabcolsep}{8pt}
\caption{Quantitative comparison of GCBM and FSCM on VITON-HD. 
The \textbf{upper} part reports the ablation results of GCBM, while the \textbf{lower} part  presents the comparison among different FSCM variants (M1--M3).}
\begin{tabular}{lcccccc}
\toprule
Method 
& FID$\downarrow$ 
& KID$\downarrow$ 
& PSNR$\uparrow$ 
& SSIM$\uparrow$ 
& LPIPS$\downarrow$ 
& DISTS$\downarrow$ \\
\midrule
\multicolumn{7}{l}{\textbf{Effectiveness of the GCBM}} \\
\midrule
w/o GCBM        
& 9.37 & \textbf{1.51} & \textbf{15.05} & 76.91 & 24.93 & 19.34 \\
\rowcolor{blue!5}
w/ GCBM (Ours)        
& \textbf{9.08} & 1.53 & 15.00 & \textbf{77.42} & \textbf{24.38} & \textbf{18.69} \\
\midrule
\multicolumn{7}{l}{\textbf{Effectiveness of the FSCM}} \\
\midrule
M1 (w/o $T_c$)
& 9.26 & 1.89 & 15.10 & 76.93 & 24.14 & 18.71 \\
M2 (w/o $F_c'$)
& 9.46 & 1.62 & \textbf{15.22} & 77.11 & \textbf{24.11} & 18.96 \\
M3 (w/o $FSCM$)
& 9.42 & 1.60 & 14.90 & 77.03 & 24.96 & 19.19 \\
\rowcolor{blue!5}
w/ FSCM (Ours)        
& \textbf{9.08} & \textbf{1.53} & 15.00 & \textbf{77.42} & 24.38 & \textbf{18.69} \\
\bottomrule
\end{tabular}
\label{tab:gcbm_fscm_comparison}
\end{table} Under the same evaluation protocol, we also report comparable results on VITON-HD.  
For qualitative comparison, all methods follow their official inference instructions, and quantitative metrics are computed uniformly from the generated images.

\noindent\textbf{{Results on the DressCode Dataset.}} Table~\ref{tab:dresscode_quant} summarizes the quantitative results on the DressCode dataset over the full test set and category-specific subsets.
Overall, BridgeDiff outperforms existing methods on the full test set and shows clear advantages on several garment categories, particularly on lower-body garments and dresses, which are more prone to side-view occlusions and structural ambiguity.
In such cases, BridgeDiff better preserves structural stability while maintaining visually continuous boundary details.
For upper-body garments, although BridgeDiff does not achieve the best scores on all perceptual metrics, it remains competitive when perceptual quality and structural consistency are jointly considered. Figure~\ref{fig:dresscode_compared} presents qualitative comparisons on the DressCode dataset.
Existing methods often suffer from structural distortions or visual discontinuities in occluded regions, whereas BridgeDiff produces flatter garment layouts with improved visual continuity.
These quantitative and qualitative improvements stem from the joint modeling of garment cues representation and explicit flat structure constraints, which stabilize garment structure during diffusion and enable continuous detail inference in occluded areas.

\noindent\textbf{{Results on the VITON-HD Dataset.}} Table~\ref{tab:vitonhd_quant} summarizes the quantitative results on the VITON-HD dataset. Overall, BridgeDiff achieves the best performance across most evaluation metrics, validating its overall advantages in perceptual quality and structural stability.
Under cross-dataset evaluation, BridgeDiff still maintains a notable performance margin, highlighting its robust generalization to unseen data distributions. Figure~\ref{fig:vitonhd_compared} presents qualitative comparisons on the VITON-HD dataset.
Existing methods often struggle to maintain geometric stability of garments and visual continuity in occluded regions, \begin{figure}[t]
    \centering
    \includegraphics[width=\textwidth,
                 trim=0.5cm 0.6cm 0.6cm 0.7cm,
                 clip]{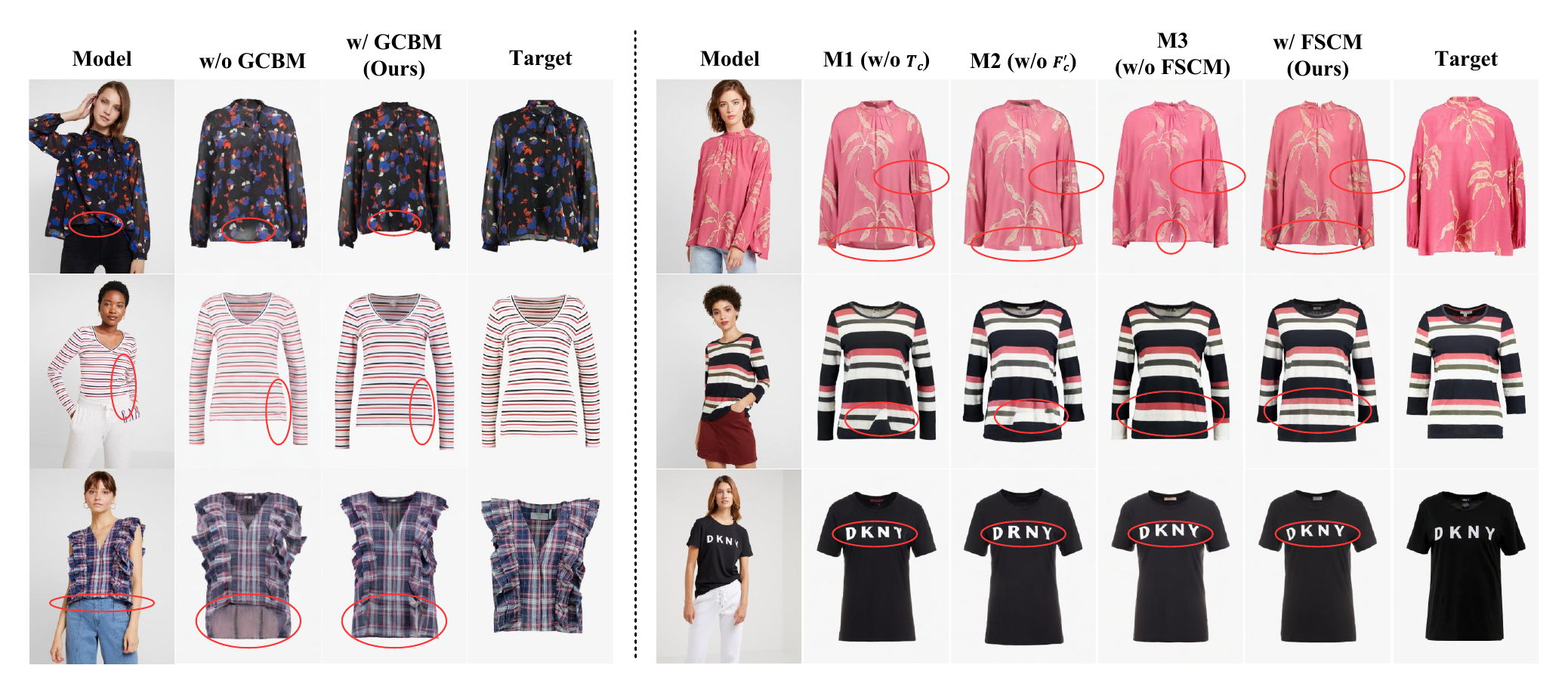}
    \caption{{Qualitative comparisons on the VITON-HD dataset.
    \textbf{Left: GCBM qualitative analysis. Right: FSCM qualitative analysis.}
    \textcolor{red}{Red circles} highlight differences in local regions across different methods.
    \textbf{Zooming in provides a clearer view of these differences.}
}
}
    \label{fig:vitonhd_ab}
\end{figure}which frequently leads to structural distortions and misaligned boundaries. In contrast, BridgeDiff is able to generate structurally stable garment results, while exhibiting superior visual continuity in occluded areas. Both quantitative and qualitative results demonstrate that incorporating explicit garment cues representation together with structured flat garment constraints facilitates the generation of garment results that are more stable and continuous in both structure and visual appearance.

\subsection{Ablation Studies and Analysis}
To analyze the effectiveness of the core components in BridgeDiff, we conduct two groups of controlled ablation studies on the VITON-HD dataset. All experiments follow the same settings as the main results unless otherwise specified. Due to space limitations, additional ablation studies and detailed analyses are provided in the \textbf{Appendix}.

\noindent\textbf{Effectiveness of the GCBM.} To verify the effectiveness of GCBM, we remove the explicit garment cues and replace them with the textual conditions from the original FSCM, while also eliminating the class embedding. The ablation variant is denoted as \textbf{w/o GCBM}, while the model equipped with explicit garment cues is denoted as \textbf{w/ GCBM}. As shown in the \textbf{upper} part of Table~\ref{tab:gcbm_fscm_comparison}, introducing GCBM leads to consistent improvements on most perceptual and structure-related metrics, indicating that the model is able to generate more continuous visual features under limited conditioning. As illustrated in the \textbf{left} part of Figure~\ref{fig:vitonhd_ab}, without GCBM, the model often produces garments with locally plausible textures but exhibits noticeable visual discontinuities in occluded or weakly observed regions. In contrast, by introducing garment cues, the model is able to naturally infer unseen regions and generate visually continuous garment features. This qualitative \begin{figure}[t]
\centering
\includegraphics[width=\textwidth,
                 trim=0.6cm 1cm 0.5cm 0.7cm,
                 clip]{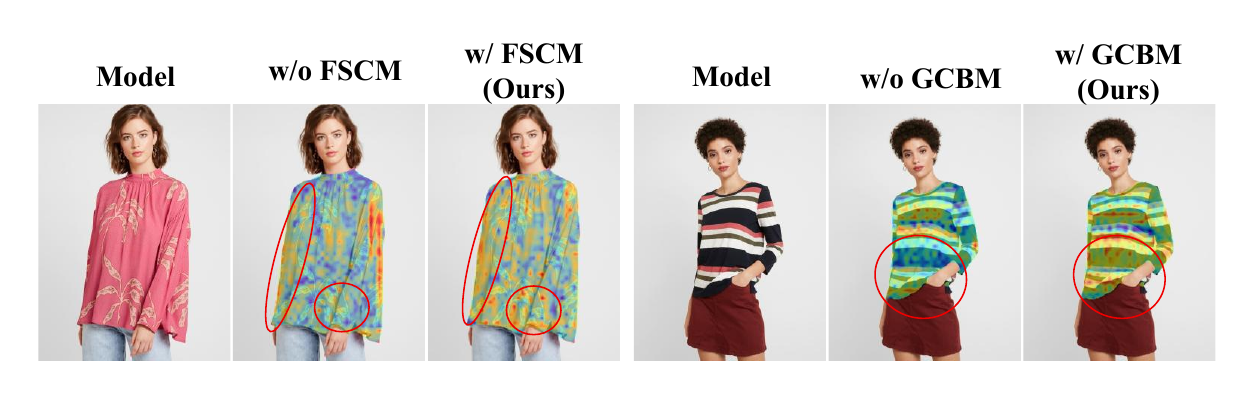}
\caption{
Heatmap visualization at the 50th denoising step on the VITON-HD dataset. 
We compare the full model with variants without GCBM or FSCM. \textcolor{red}{Red circles} highlight differences in local regions across different methods.
}
\label{fig:attention_vis}
\end{figure}comparison highlights the role of GCBM in generating garment cues, which effectively improves the continuity of appearance modeling under limited or partial observations.

\noindent\textbf{{Effectiveness of the FSCM.}} To evaluate the effectiveness of FSCM, we design three ablation variants while keeping all other components unchanged. 
Specifically, \textbf{M1} removes $T_c$ and conditions the model only on the image-encoder-based garment features $F_c'$, while \textbf{M2} removes $F_c'$ and retains only the flat garment textual description $T_c$ as the structural condition. 
These two variants evaluate whether a single garment condition is sufficient to stabilize garment structure. In \textbf{M3}, the FSCM module is entirely removed and replaced by a prompt-level augmentation, where additional flat garment descriptions $T_c'$ are prepended to the original garment appearance prompt\footnote{The added textual description mainly specifies key flat garment appearance attributes. Concretely, we prepend ``A flat-lay \texttt{<category>},'' to the original garment appearance prompt for all target flat garment images.}. 
This setting assesses whether such naive prompt-level augmentation can maintain garment structural stability while preserving the correctness and continuity of visual information. As shown in the \textbf{lower} part of Table~\ref{tab:gcbm_fscm_comparison}, the complete FSCM demonstrates the best overall performance, leading across multiple structure-related metrics, which indicates its superior effectiveness in maintaining the global structural stability of flat garments. Although its PSNR and LPIPS are slightly lower than those of some variants, this is expected, as the flat-structure constraint module emphasizes canonical flat garment geometry rather than pixel-level similarity or local perceptual consistency. At the same time, its advantages in distribution-level metrics such as FID further indicate that the use of the complete FSCM enables more stable preservation of the overall garment structure compared to other variants. 
As shown in the \textbf{right} part of Fig.~\ref{fig:vitonhd_ab}, relying on a single garment condition often leads to local wrinkles, structural errors, or discontinuous visual features. 
Even M3, which uses prompt-level augmentation, cannot consistently maintain structural stability and still produces subtle wrinkles, geometric inconsistencies, or incorrect fine-grained details.
In contrast, FSCM leverages complementary visual cues and explicit textual constraints to ensure structural stability in flat garments and generate continuous visual results.

\noindent\textbf{Attention Heatmap Visualization.} To analyze the effects of the proposed modules, we visualize the attention weights in the denoising UNet at the 50th denoising step. 
Specifically, we extract the attention distribution from \begin{figure}[t]
    \centering
    \includegraphics[width=\textwidth,
                 trim=0.5cm 0.6cm 0.5cm 0.6cm,
                 clip]{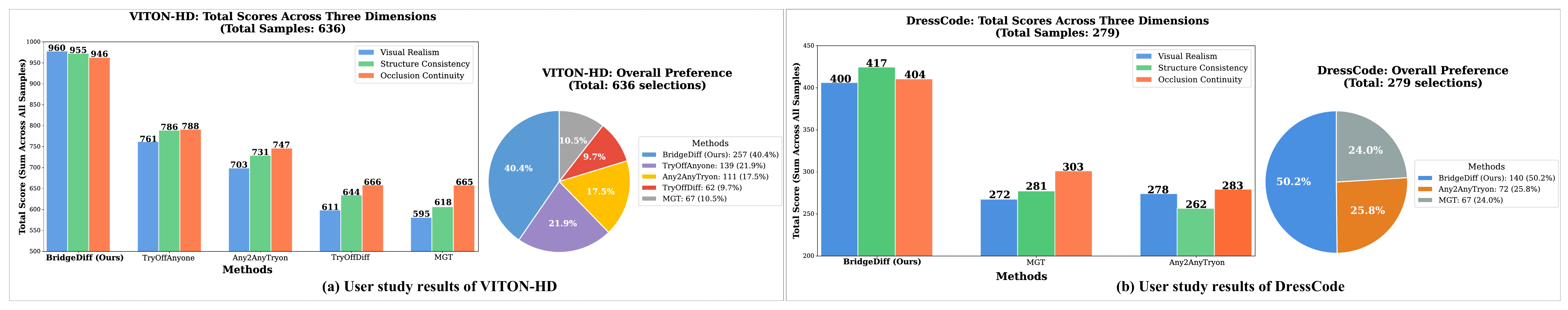}
    \caption{
    Summary of the user study results.
(a) VITON-HD. (b) DressCode.
Within each dataset, the left panel shows the average perceptual scores of different methods across three dimensions as bar charts, while the right panel presents the overall user preference distribution as pie charts. 
    }
    \label{fig:user_results}
\end{figure}the down blocks.0.attentions.1 layer, where each spatial position of the denoising latent attends to the target garment features. The aggregated attention weights are mapped into spatial heatmaps for visualization. As shown in Fig.~\ref{fig:attention_vis}, the full model with FSCM assigns higher 
attention weights around garment structural boundaries, indicating improved 
modeling of flat garment layouts. In contrast, the variant without FSCM mainly 
focuses on locally visible regions. For GCBM, the model maintains strong 
attention responses in occluded areas, while removing GCBM leads to noticeably 
weaker attention in these regions.
\subsection{User Study}
To evaluate the perceptual quality and structural consistency of generated garments, 
we conduct a questionnaire-based user study. Participants are shown anonymized and 
shuffled results from different methods under the same input model image, with 
multiple random seeds for fairness. Each result is evaluated based on 
\textbf{Visual Realism}, \textbf{Garment Structure Consistency}, 
\textbf{Occluded Region Continuity}, and \textbf{Overall Preference}. 
The first three criteria are rated on a three-point scale (0--2), while overall 
preference is collected as a single-choice selection. Results are aggregated by 
the total number of evaluated samples. As shown in Fig.~\ref{fig:user_results}, 
our method achieves consistently higher ratings on both VITON-HD and DressCode, 
with more participants selecting our results as their preferred outputs.

\section{Conclusion and Future Work}
We proposed BridgeDiff, which explicitly bridges the gap between human-centric garment observations and canonical flat garment synthesis. By introducing garment cue representations and injecting explicit flat-structure constraints during denoising, BridgeDiff effectively alleviates visual discontinuities in unobservable regions and improves structural stability compared to text-only constrained approaches. Extensive experimental results demonstrate that BridgeDiff achieves competitive performance across multiple garment categories, achieving better visual continuity in occluded regions and enhanced overall structural stability. Nevertheless, the method can still be imprecise or incomplete under severe occlusions or extreme poses. In future work, we plan to train on larger datasets to further improve the model's ability to infer visually continuous garment regions.

%
%
\bibliographystyle{splncs04}
\bibliography{main}

\clearpage
\appendix

\begin{center}
{\LARGE \textbf{Appendix}}
\end{center}

\section*{Supplementary Material}
\addcontentsline{toc}{section}{Supplementary Material}
This supplementary material provides an extended and in-depth exploration of the experiments and methodologies presented in the main paper. Section~\ref{sec:prelim} introduces additional background and defines key symbols to facilitate a clearer understanding of the proposed framework. 
Section~\ref{sec:train_infer} describes the training strategy and inference pipeline in detail, covering important implementation choices and practical considerations. 
Section~\ref{sec:add_exp} presents extended experimental analyses, including more comprehensive ablation studies. 
Section~\ref{sec:user_study} provides a detailed description of the user study procedure.
Section~\ref{sec:text_gen} explains the procedure used to generate garment-related textual descriptions. 
Section~\ref{sec:add_qual} showcases a broader range of qualitative comparison examples with state-of-the-art methods, offering further insights into the strengths of our approach.
\section{Preliminaries and Notations}
\label{sec:prelim}
\subsection{Preliminaries}
\subsubsection{Latent Diffusion Models.} The core idea of Latent Diffusion Models (LDMs) \cite{rombach2022high} reduce the computational cost of diffusion-based generative models by conducting the diffusion process in a compact latent space. 
Given an input image $x_0$, a pre-trained Variational Autoencoder (VAE) \cite{kingma2013auto} is first used to encode the image into a latent representation, enabling efficient training and inference while maintaining high generation quality. An LDM mainly consists of a denoising UNet $\epsilon_\theta(\cdot, t)$ and a VAE composed of an encoder $\mathcal{E}$ and a decoder $\mathcal{D}$. During training, Gaussian noise is progressively added to the clean data according to the forward diffusion process, producing a noisy sample at timestep $t$:
\begin{equation}
z_t = \sqrt{\bar{\alpha}_t} \, z_0 + \sqrt{1 - \bar{\alpha}_t} \, \epsilon,
\end{equation}
where $z_0 = \mathcal{E}(x_0)$ denotes the latent representation of the input image $x_0$ obtained from the pre-trained VAE encoder $\mathcal{E}$, $\epsilon \sim \mathcal{N}(0,1)$ is standard Gaussian noise, and $\bar{\alpha}_t$ is a predefined function of $t$ that determines the diffusion schedule.
The denoising UNet is trained to predict the injected noise from the noisy input $z_t$, and the training objective is defined as:
\begin{equation}
\mathcal{L}_{\mathrm{LDM}} := 
\mathbb{E}_{z_0, \epsilon, t} \left[ \left\| \epsilon - \epsilon_\theta(z_t, t) \right\|_2^2 \right]
\end{equation}\begin{table}[h]
\centering
\caption{Notation Definition}
\begin{tabular}{l|l}  
\hline
\textbf{Symbol} & \textbf{Definition} \\
\hline
$X_m$ & Model image \\
$X_w$ & Warped garment \\
$X_c$ & Target garment image \\
$F_w$ & Encoded feature of warped garment \\
$F_m$ & Encoded feature of model image (stage 1) \\
$F_c$ & Output of garment condition bridge module (GCBM) \\
$\epsilon$ & Gaussian noise \\
$t$ & Timestep \\
$Z_m$ & Model image encoded via VAE \\
$I_m$ & Encoded feature of model image (stage 2) \\
$I'_m$ & Projected $I_m$ \\
$T_m$ & Encoded textual description of model \\
$M_c$ & Cloth-agnostic mask \\
$M_o$ & Intermediate features of model UNet \\
$M'_o$ & Features injected into denoising UNet \\
$Z_c$ & Target garment encoded via VAE \\
$z_c$ & $Z_c$ with added noise \\
$T_{flat}$ & Encoded textual description of flat garment \\
$F'_c$ & Projected output of garment condition bridge module (GCBM) \\
$G$ & Number of attention heads in flat structure constraint module (FSCM) \\
$Z_{fc}$ & Final output of flat structure constraint module (FSCM) \\
$T_c$ & Encoded appearance description of garment \\
$\hat{z}_c$ & Rectified input of target garment \\
$\hat{\epsilon}$ & Rectified new target noise \\
$\lambda$ & Balancing hyperparameter for model UNet hybrid cross-attention \\
$\beta$ & Balancing hyperparameter for denoising UNet hybrid attention \\
$\gamma$ & Balancing hyperparameter for denoising UNet hybrid cross-attention \\
$w$ & Classifier-Free Guidance (CFG) weight \\
\hline
\end{tabular}
\label{tab:notation}
\end{table}where $t \in \{1, \dots, T\}$ denotes the timestep of the diffusion process. 
At inference time, samples drawn from the latent distribution are iteratively denoised and finally mapped back to the image space through a single forward pass of the VAE decoder $\mathcal{D}$.

\subsubsection{Diffusion Rectification and Estimation-Adaptive Models.} 
\label{DREAM}
Diffusion Rectification and Estimation-Adaptive Models (DREAM) \cite{zhou2024dream} is a training strategy designed to alleviate the trade-off between perceptual quality and pixel-level distortion in conditional generation tasks.  
During training, the diffusion model computes a rectified input and a corresponding target as follows:
\begin{equation}
\hat{z}_t = z_t + \sqrt{1 - \bar{\alpha}_t} \, \lambda_t \, \Delta \epsilon_{t,\theta}, \quad
\hat{\epsilon} = \epsilon + \lambda_t \, \Delta \epsilon_{t,\theta},
\end{equation}
where $\Delta \epsilon_{t,\theta} = \epsilon - \mathrm{StopGradient}(\epsilon_\theta(z_t, t))$, $\lambda_t = (\sqrt{1 - \bar{\alpha}_t})^p$ with $p$ being a hyperparameter, $\hat{z}_t$ denotes the rectified input and $\hat{\epsilon}$ is the new target. 
The training objective is defined as:
\begin{equation}
\mathcal{L}_{\mathrm{DREAM}} := 
\mathbb{E}_{z_t, \hat{\epsilon}, t} \Big[ \big\| \hat{\epsilon} - \epsilon_\theta(\hat{z}_t, t) \big\|_2^2 \Big],
\end{equation}
During inference, the standard reverse diffusion process \cite{song2020denoising} is still adopted. DREAM improves training accuracy and convergence efficiency, although each training step requires an additional forward pass, slightly increasing the computational cost.
\subsection{Notations}
All symbols and their corresponding definitions are provided in detail in Table~\ref{tab:notation}.

\section{Training and Inference Details}
\label{sec:train_infer}
The proposed Flat Structure Constraint for Conditional Diffusion is optimized using only a single training objective, without introducing any auxiliary losses. Specifically, for the model UNet, all parameters as well as the image projection layers are fully trainable. In contrast, for the denoising UNet, we only update the class embedding layer, all linear projection layers within the FSCM, the key and value projection matrices $W_k^f$ and $W_v^f$ at the specific $j$-th cross-attention layer, and the projection matrices $\hat{W}_k$ and $\hat{W}_v$ in all hybrid attention layers. All remaining parameters are kept frozen during training.

The training objective follows the standard diffusion noise learning formulation. Given the model features $I_m'$, model text descriptions $T_m$, clean model latent representations $Z_m$, garment appearance text embeddings $T_c$, garment cues representation $F_c$, projected garment cues representation $F_c'$, flat garment text embeddings $T_{flat}$, Gaussian noise $\epsilon \sim \mathcal{N}(0,1)$, added noisy latent $z_c$ and diffusion timestep $t$, the diffusion loss is defined as:
\begin{equation}
\begin{split}
\mathcal{L}_{\mathrm{diff}} &= 
\mathbb{E}_{I_m', T_m, Z_m, T_c, F_c, F_c', T_{flat}, \epsilon, z_c,t} 
\Big[ 
\big\| 
\hat{\epsilon} - \epsilon_\theta(
\hat{z_c},\\
&\quad I_m', T_m, Z_m, T_c, F_c, F_c', T_{flat}, t)
\big\|_2^2
\Big],
\end{split}
\end{equation}
where $\epsilon_\theta(\cdot)$ denotes the noise prediction network, $\hat{z_c}$ denoted the rectified input and $\hat{\epsilon}$ is the new target \cite{zhou2024dream}. 

During inference, we adopt Classifier-Free Guidance (CFG)\cite{ho2022classifier} to perform conditional sampling. Specifically, given the noisy latent $\epsilon$, the conditional noise prediction $\epsilon_\theta(\epsilon, I_m', T_m, Z_m, T_c,F_c, F_c', T_{flat}, t)$ and the unconditional noise prediction $\epsilon_\theta(\epsilon,t)$, the final guided noise estimate is obtained as:

\begin{equation}
\hat{\epsilon}_o = \epsilon_\theta(\epsilon, t) + w \Big( \epsilon_\theta(\epsilon, I_m', T_m, Z_m, T_c, F_c, F_c', T_{flat}, t) - \epsilon_\theta(\epsilon, t) \Big)
\end{equation}
where $w$, often referred to as the guidance scale is a scalar that controls the strength of conditioning.

\begin{table}[t]
\centering
\small
\setlength{\tabcolsep}{9pt}
\caption{Ablation study on FSCM insertion positions on the VITON-HD dataset. $\uparrow$ indicates higher is better, $\downarrow$ indicates lower is better. \textbf{Bold} denotes the best result.}
\begin{tabular}{lcccccc}
\toprule
Method & FID$\downarrow$ & KID$\downarrow$ & PSNR$\uparrow$ & SSIM$\uparrow$ & LPIPS$\downarrow$ & DISTS$\downarrow$ \\
\midrule
Down 1 & 9.42 & 1.86 & 15.00 & \textbf{77.49} & \textbf{24.18} & 18.82 \\
Down 2 & 9.32 & 1.88 & 15.03 & 77.24 & 24.27 & 18.77 \\
Mid    & 9.25 & 1.84 & \textbf{15.07} & 77.36 & 24.22 & 18.77 \\
Up 1   & 9.42 & 1.81 & 15.03 & 77.48 & 24.20 & 18.73 \\
Up 2   & \textbf{9.03} & 1.67 & 15.05 & 77.00 & 24.42 & 18.71 \\
Up 3   & 9.35 & 1.85 & 14.92 & 77.06 & 24.37 & 18.71 \\
All    & 9.51 & 1.91 & 14.84 & 77.09 & 24.53 & 18.81 \\
\rowcolor{blue!5}
Down 0 (Ours) & 9.08 & \textbf{1.53} & 15.00 & 77.42 & 24.38 & \textbf{18.69} \\
\bottomrule
\end{tabular}
\label{tab:fscm_pos}
\end{table}

\section{Additional Experimental Results and Analysis}
\label{sec:add_exp}
To analyze the effectiveness of the core components in BridgeDiff, we conduct multiple controlled ablation and comparative studies on the VITON-HD dataset.  
All experiments follow the same settings as the main paper unless otherwise noted. Specifically, we investigate the impact of the FSCM insertion position on generation quality (see Section~\ref{sec:exp_fscm_pos}), compare BridgeDiff with UniFit under the virtual try-off (VTOFF) setting to analyze how different methods address the challenge of focusing on garment regions (see Section~\ref{sec:exp_attention}), and analyze the effect of removing the model textual description $T_m$ and garment appearance textual description $T_c$ (see Section~\ref{sec:exp_text}).
Finally, we investigate the sensitivity of key hyperparameters of the proposed framework (see Section~\ref{sec:exp_hyperparam}).
Taken together, these analyses systematically demonstrate the rationality and effectiveness of the key design choices in BridgeDiff.
\subsection{FSCM Insertion Position Analysis}
\label{sec:exp_fscm_pos}
This subsection investigates how the insertion point of the FSCM affects the stability and visual quality of generated flat garments. 
In our experiments, we leverage the layer naming convention from \texttt{unet.attn\_processors} to denote the FSCM insertion positions. 
Specifically, \textbf{Down 0, Down 1, Down 2} correspond to the three consecutive downsampling stages of the UNet, while \textbf{Up 1, Up 2, Up 3} indicate the respective upsampling stages. The \textbf{Mid} stage refers to the intermediate bottleneck layer connecting the encoder and decoder. 
In addition, the variant \textbf{All} refers to inserting FSCM into all available stages simultaneously. This setup allows us to systematically analyze the effect of structural guidance on global garment geometry and visual continuity, providing insights into where explicit flat-structure constraints are most effective. 
\begin{figure}[t]
    \centering
    \includegraphics[width=\textwidth,
                 trim=0.5cm 0.6cm 0.6cm 0.7cm,
                 clip]{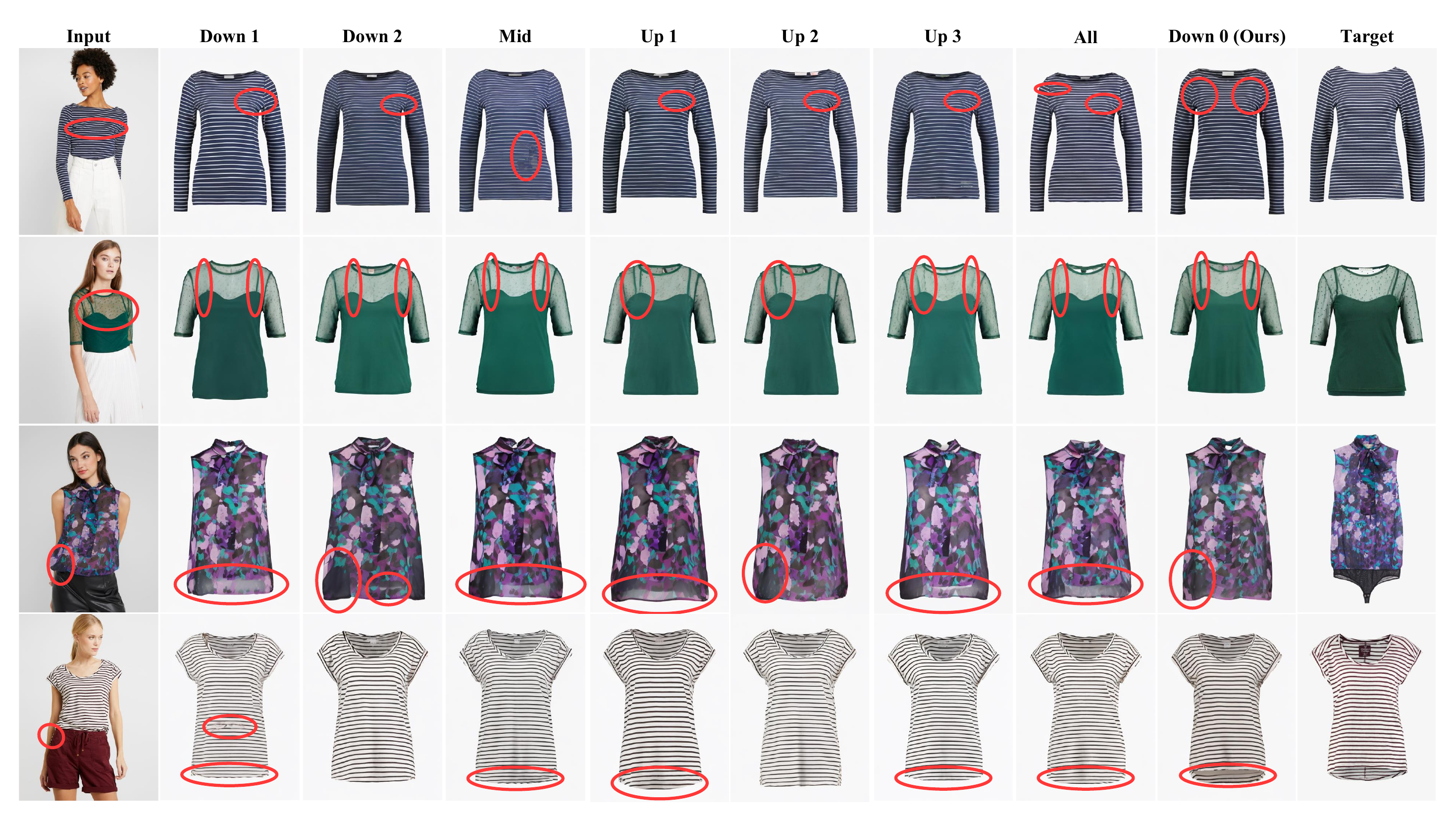}
    \caption{Qualitative comparisons of FSCM Insertion Position on the VITON-HD dataset.
    \textcolor{red}{Red circles} highlight differences in local regions across different settings. 
    \textbf{Zooming in provides a clearer view of these differences.}}
    \label{fig:ipaweizhi}
\end{figure}

As shown in Table~\ref{tab:fscm_pos}, \textbf{Down 0} achieves the most balanced performance across distribution-level, perceptual, and low-level reconstruction metrics.
Specifically, \textbf{Down 0} yields the lowest KID and DISTS, while maintaining competitive FID and SSIM, indicating improved global structural stability and visual continuity. In comparison, inserting FSCM at a slightly later downsampling stage (e.g., \textbf{Down 1}) achieves relatively strong performance on SSIM and LPIPS, but does not provide clear advantages on distribution-level metrics (e.g., FID and KID), suggesting that structural constraints introduced at this stage are less effective in stabilizing global garment structure.
Furthermore, inserting FSCM at later stages (e.g., \textbf{Up 2}) slightly improves FID and PSNR, but leads to inferior performance on perceptual or structure-related metrics, indicating less consistent garment structure.
Similarly, inserting FSCM into all layers results in degraded performance across most metrics, implying that overly strong structural constraints may hinder effective modeling during the diffusion process.
Overall, these results demonstrate that introducing flat structure constraints at the earliest stages of the diffusion process is more effective for stabilizing flat-garment geometry while preserving overall visual quality. Qualitative comparisons in Fig.~\ref{fig:ipaweizhi} further highlight the impact of FSCM insertion positions.
In the first three examples, alternative insertion strategies frequently introduce unnatural wrinkles, fragmented textures, or visually discontinuous regions, especially around areas with complex folds, hair occlusions, or garment boundaries.
Even when the overall layout appears plausible, these variants often fail to maintain continuous and coherent visual information under partial observations. More interestingly, the fourth example reveals a subtle yet informative case.
Although only a small and easily overlooked garment region is visible near the model’s lower-left boundary (i.e., the upper garment region highlighted by the red circle), most insertion variants are able to capture this weak visual cue and generate different hem configurations.
However, \textbf{only our early insertion strategy} produces a result that is closest to the ground truth, correctly inferring the garment structure while maintaining visual continuity.
This observation further demonstrates that GCBM can effectively generate garment cue representations and interact with the flat textual representation at an early stage. However, incorporating critical structural information before geometric ambiguities accumulate during the diffusion process is particularly important for maintaining consistent garment structure.

\subsection{Comparison with UniFit's Spatial Attention Focusing Loss}
\label{sec:exp_attention}
In VTOFF tasks, diffusion models often attend to regions in the model image that are irrelevant to the target garment (e.g.,background or non-garment body parts), which degrades fine-grained garment modeling and introduces visual artifacts. To address this issue, UniFit~\cite{zhang2025unifit} is inspired by DreamO~\cite{mou2025dreamo} and introduces a spatial attention focusing loss that explicitly regularizes cross-attention maps to encourage attention concentration on garment regions. Specifically, UniFit aggregates cross-attention responses between denoising queries and reference image (i.e., model image) keys across attention heads and reference tokens. Finally, a spatial attention map $M$ is aligned with a pre-extracted cloth-agnostic mask $M_c$ from the model image via an MSE loss:
\begin{equation}
\mathcal{L}_{\text{focus}} =
\frac{1}{N_R N_L}
\sum_{j=1}^{N_L}
\sum_{i=1}^{N_R}
\left\| M_i^j - M_{c,i} \right\|_2^2 .
\end{equation}where $M_i^j$ denotes the response map for the $i$-th model image in the $j$-th attention layer, $M_{c,i}$ is the cloth-agnostic mask corresponding to the $i$-th model image, and $N_R$ and $N_L$ represent the number of model images and attention layers, respectively. In contrast, we adopt a simpler and more efficient strategy. Instead of introducing an additional optimization objective, we directly leverage the cloth-agnostic mask to extract clean garment regions from the model image before attention computation. This design avoids interference from irrelevant regions without extra training losses, leading to a simpler and more direct optimization process. In our experiments, we compare our method with UniFit and denote the variant with the attention focusing loss as \textbf{w/ Loss}, and the variant that directly extracts garment regions as \textbf{w/o Loss}.
\begin{table}[t]
\centering
\setlength{\tabcolsep}{8pt}
\caption{Ablation study results on the VITON-HD dataset.
$\uparrow$ indicates higher is better, $\downarrow$ indicates lower is better.
\textbf{Bold} denotes the best result.
\textbf{The upper part} reports the comparison with UniFit’s spatial attention focusing loss,
while \textbf{the lower part} presents the impact of textual descriptions.}
\begin{tabular}{lcccccc}
\toprule
Method 
& FID$\downarrow$ 
& KID$\downarrow$ 
& PSNR$\uparrow$ 
& SSIM$\uparrow$ 
& LPIPS$\downarrow$ 
& DISTS$\downarrow$ \\
\midrule

\multicolumn{7}{l}{\textbf{Comparison with UniFit’s Spatial Attention Focusing Loss}} \\
\midrule
w/ Loss (UniFit)        
& 9.78 & 1.99 & 14.92 & 77.37 & 24.67 & \textbf{18.66} \\
\rowcolor{blue!5}
w/o Loss (Ours)        
& \textbf{9.08} & \textbf{1.53} & \textbf{15.00} & \textbf{77.42} & \textbf{24.38} & 18.69 \\
\midrule

\multicolumn{7}{l}{\textbf{Impact of Textual Descriptions}} \\
\midrule
w/o Text         
& \textbf{9.06} & 1.57 & \textbf{15.00} & 77.07 & 24.42 & 18.76 \\
\rowcolor{blue!5}
w/ Text (Ours)        
& 9.08 & \textbf{1.53} & \textbf{15.00} & \textbf{77.42} & \textbf{24.38} & \textbf{18.69} \\
\bottomrule
\end{tabular}
\label{tab:ablation_combined}
\end{table}

As shown in the \textbf{upper} part of Table~\ref{tab:ablation_combined}, directly extracting garment regions consistently outperforms the UniFit variant using the spatial attention focusing loss across nearly all metrics.
This indicates that directly extracting garment regions via the cloth-agnostic mask is more effective than relying on attention regularization for suppressing irrelevant information, leading to cleaner garment reconstruction and improved perceptual and structural fidelity. Qualitative comparisons in Figure~\ref{fig:loss_ablation_qual} further illustrate this effect. When the model wears semi-transparent garments, BridgeDiff is able to accurately generate the clean underlying garment, whereas UniFit tends to retain colors from the visible body. Additionally, some visual information around the garment is also incorporated into the generated clothing.
We believe that since the cross-attention uses $\mathrm{softmax}$, non-garment regions still receive non-zero attention scores, leading to extra details from irrelevant areas at the garment. In contrast, directly masking out irrelevant regions ensures the generated garments focus solely on the target cloth, improving both visual clarity and structural consistency. 
\textbf{It is worth noting that our direct garment region extraction is extremely simple and effective, and more importantly, it does not introduce any additional training objectives.}

\subsection{Impact of Textual Descriptions}
\label{sec:exp_text}
To study the role of textual descriptions in our model, we analyze the impact of removing the model textual description $T_m$ and the garment appearance textual description $T_c$ on performance. Specifically, for the ablation study, we define the variant without textual inputs as \textbf{w/o Text}. In this variant, the model UNet receives the encoded model image, which is first processed by the image encoder and then projected through a linear layer before being injected into the UNet. This injection procedure follows the same design as the IP-Adapter \cite{ye2023ip}. Meanwhile, the denoising UNet no longer receives the garment appearance textual description $T_c$. Instead, only the FSCM is injected into the cross-attention layers, while all other cross-attention inputs are replaced with empty text tokens. In contrast, the variant using both the model textual description $T_m$ and the garment appearance textual description $T_c$ is defined as \textbf{w/ Text}.

\begin{figure}[t]
  \centering
  \begin{subfigure}{0.48\linewidth}
    \includegraphics[width=\textwidth,
                 trim=0.5cm 0.6cm 0.6cm 0.7cm,
                 clip]{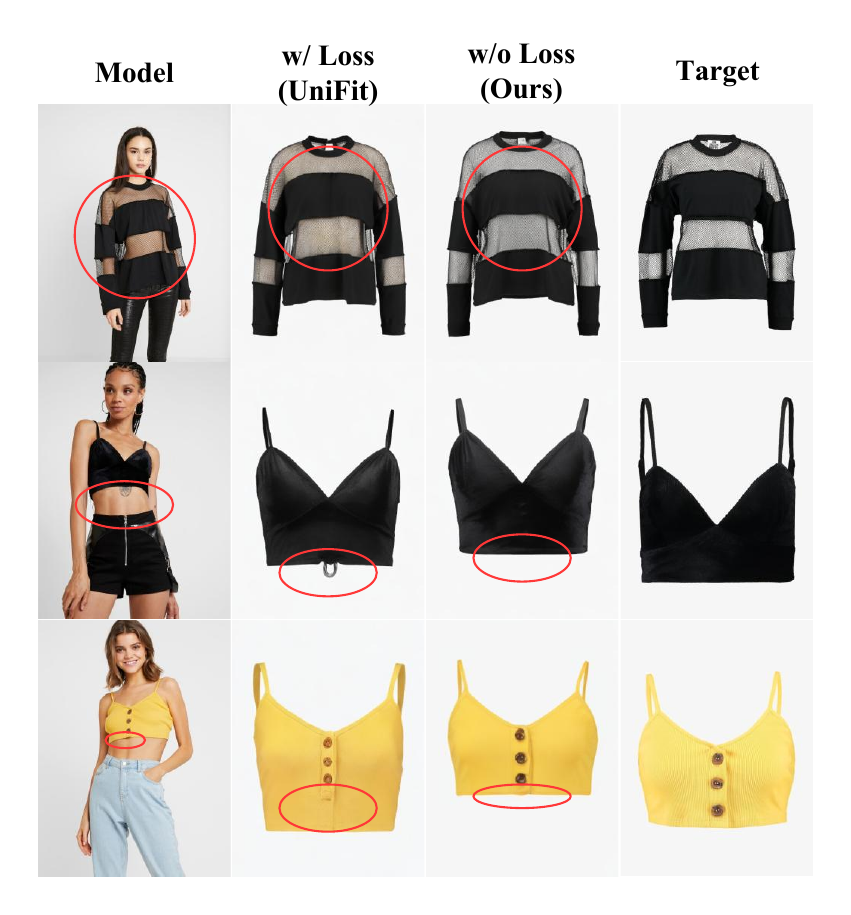}
    \caption{Qualitative comparison of UniFit attention loss vs. our direct garment region extraction.}
    \label{fig:loss_ablation_qual}
  \end{subfigure}
  \hfill
  \begin{subfigure}{0.48\linewidth}
    \includegraphics[width=\textwidth,
                 trim=0.5cm 0.6cm 0.6cm 0.7cm,
                 clip]{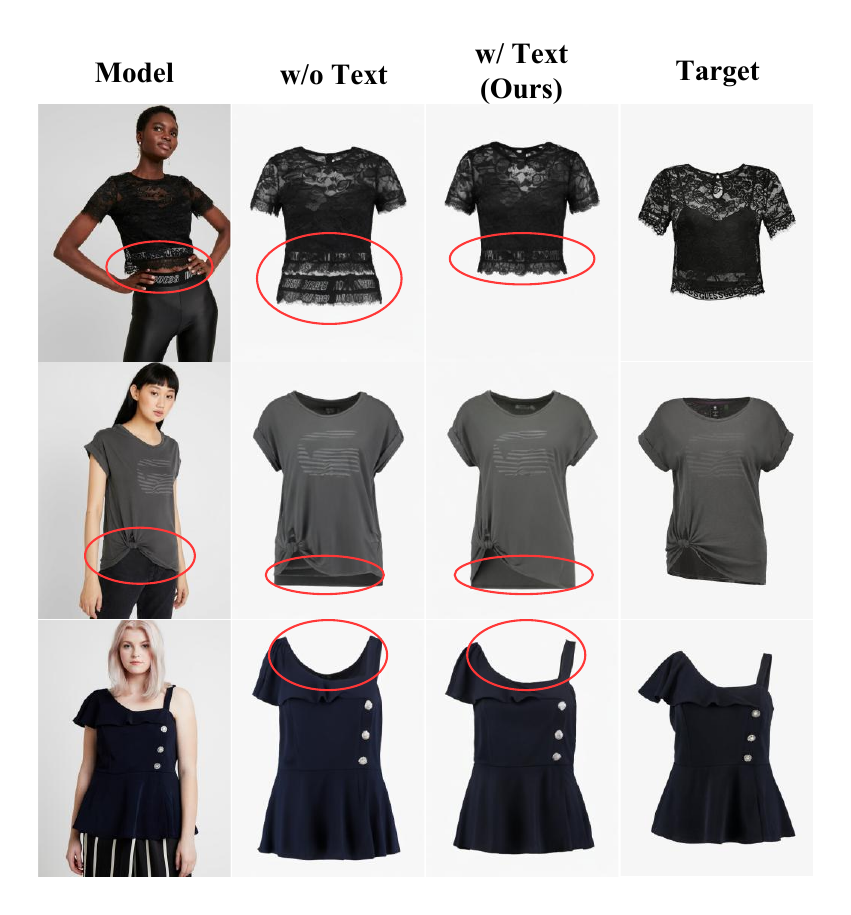}
    \caption{Qualitative comparison illustrating the impact of textual descriptions.}
    \label{fig:text_ablation_qual}
  \end{subfigure}
  \caption{Qualitative comparisons on the VITON-HD dataset.
  \textcolor{red}{Red circles} highlight differences in local regions across different settings.
  \textbf{Zooming in provides a clearer view of these differences.}}
  \label{fig:qual_ablation}
\end{figure}

As shown in the \textbf{lower} part of Table~\ref{tab:ablation_combined}, removing both the model textual description $T_m$ and the garment appearance description $T_c$ does not lead to a significant degradation in overall quantitative performance. This surprising result suggests that the core structural and visual information required for garment synthesis are largely captured by the FSCM and the garment cues, rather than relying heavily on textual guidance. Nevertheless, the noticeable drop in SSIM indicates that textual descriptions still contribute to improving structural consistency in specific regions (e.g., garment necklines or waists). Qualitative comparisons in Figure~\ref{fig:text_ablation_qual} further illustrate the role of textual descriptions. Without text guidance, the model may introduce visually plausible but semantically irrelevant details in ambiguous garment regions, especially near boundaries areas. In contrast, incorporating textual descriptions encourages the model to focus on the currently visible garment semantics, leading to more coherent and semantically consistent synthesis results.

 \begin{figure}[t]
    \centering
    \includegraphics[width=\textwidth,
                 trim=0.5cm 0.6cm 0.6cm 0.7cm,
                 clip]{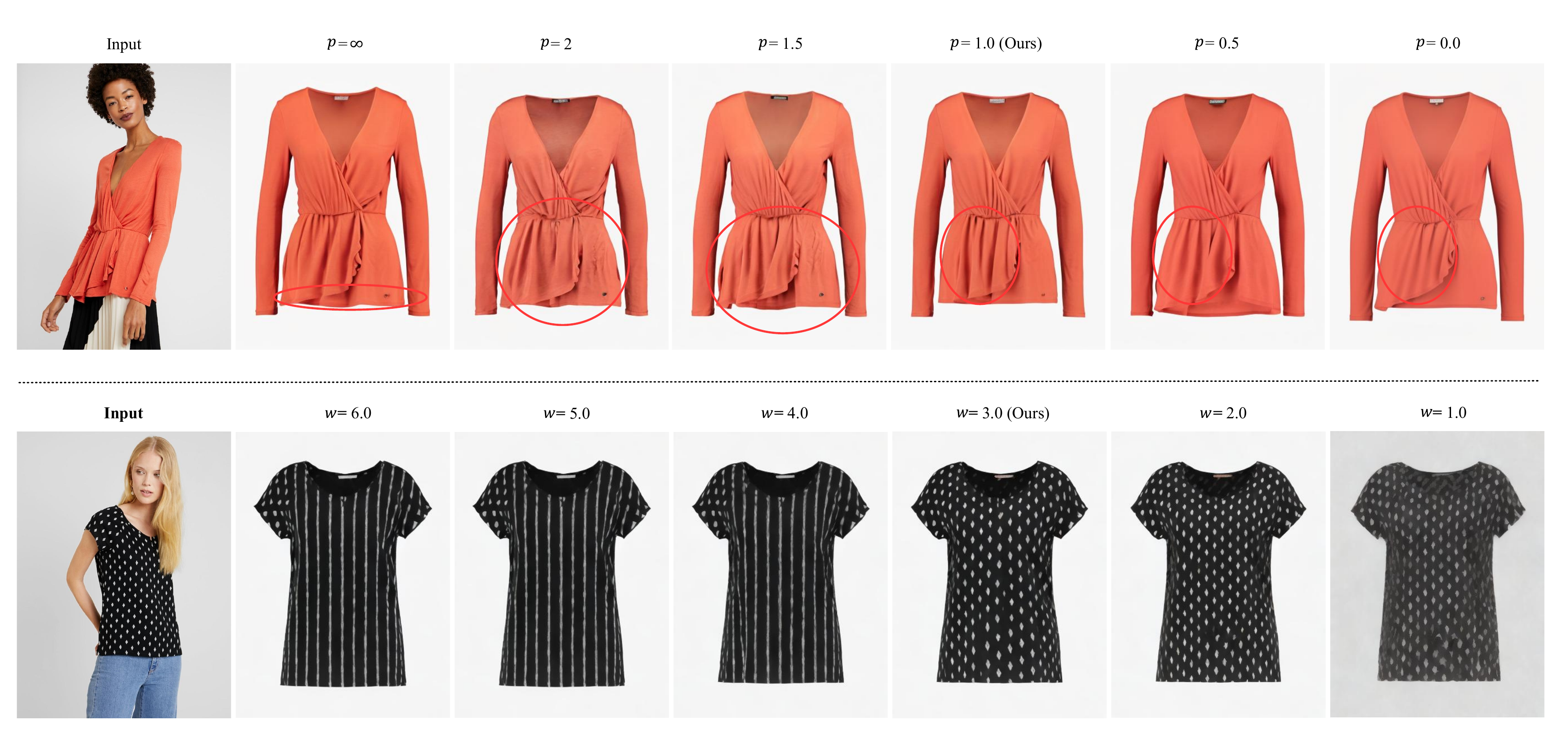}
    \caption{{Qualitative comparisons on the VITON-HD dataset.
    \textbf{Upper: DREAM qualitative analysis. Lower: CFG qualitative analysis.}
    \textcolor{red}{Red circles} highlight differences in local regions across different methods.
    \textbf{Zooming in provides a clearer view of these differences.}
}}
    \label{fig:dream_cfg}
\end{figure}
\subsection{Hyperparameter Sensitivity}
\label{sec:exp_hyperparam}
We validate two key hyperparameters in our method. Specifically, we investigate the effects of $p$, the hyperparameter in DREAM \cite{zhou2024dream}, which is designed to alleviate the trade-off between perceptual quality and pixel-level distortion in conditional generation tasks, and $w$, the CFG \cite{ho2022classifier} weight that balances conditional faithfulness and sample diversity during generation.

\subsubsection{Diffusion Rectification and Estimation-Adaptive Models.} As described in Sec.~\ref{DREAM}, we incorporate DREAM to improve the training of conditional diffusion models. The hyperparameter $p$ controls the strength of DREAM. Specifically, DREAM is disabled and the training objective reduces to the standard diffusion formulation when $p = \infty$, whereas DREAM is activated when $p$ takes finite values.

We qualitatively analyze the effect of DREAM. As shown in the \textbf{upper} part of Figure~\ref{fig:dream_cfg}, a small $p$ leads to overly smooth outputs, while a large $p$ introduces excessive high-frequency noise, resulting in noticeable wrinkles and local irregularities in the garments. This highlights the importance of properly selecting the strength of DREAM to balance perceptual fidelity and distortion regulation. Table~\ref{tab:dream_lambda} presents the results of training with different $p$ values on the VITON-HD dataset. As $p$ increases, distribution-level metrics (such as FID and KID) initially improve but increase again at larger $p$ values, while LPIPS exhibits a similar trend. Overall, $p = 1.0$ achieves a good balance across all metrics.

\subsubsection{Classifier-Free Guidance.}
CFG balances between conditional and unconditional generation,
guiding the model to produce results that better satisfy the conditioning constraints
while maintaining higher visual fidelity.
To evaluate the effect of CFG on generation quality,
we conduct inference strengths of 1.0, 2.0, 3.0, 4.0, and 5.0,
while keeping all other experimental settings fixed,
where $w$ = 1.0 corresponds to standard conditional generation.

As shown in the \textbf{lower} part of Figure~\ref{fig:dream_cfg}, increasing the CFG strength improves image detail and fidelity. However, when the strength exceeds 3.0, the generated images exhibit severe artifacts and high-frequency noise, resulting in degraded visual quality. In our experiments, we found that a CFG strength between 2.0 and 3.0 produces the most realistic and natural results. Therefore, we set the CFG strength to 3.0 for all experiments. 
\begin{table}[t]
\centering
\small
\setlength{\tabcolsep}{10pt}
\caption{Ablation results of different $p$ in DREAM on the VITON-HD dataset. $p = \infty$ means DREAM is disabled. $\uparrow$ indicates higher is better, $\downarrow$ indicates lower is better. \textbf{Bold} denotes the best result.}
\begin{tabular}{lcccccc}
\toprule
$p$ & FID$\downarrow$ & KID$\downarrow$ & PSNR$\uparrow$ & SSIM$\uparrow$ & LPIPS$\downarrow$ & DISTS$\downarrow$ \\
\midrule
0.0        & 11.92 & 3.58 & 14.99 & \textbf{77.58} & 25.09 & 20.51 \\
0.5      & 9.60 & 2.04 & 14.96 & 77.41 & 24.60 & 19.14 \\
\rowcolor{blue!5} 1.0 (Ours) & \textbf{9.08} & \textbf{1.53} & \textbf{15.00} & 77.42 & \textbf{24.38} & 18.69 \\
1.5      & 9.13 & 1.58 & 14.89 & 76.86 & 24.51 & 18.66 \\
2.0      & 9.42 & 1.84 & 14.85 & 76.66 & 24.44 & 18.75 \\
$\infty$ & 9.15 & 1.63 & 14.87 & 77.28 & 24.64 & \textbf{18.66} \\
\bottomrule
\end{tabular}
\label{tab:dream_lambda}
\end{table}

\section{User Study Example Questionnaire}
\label{sec:user_study}
\begin{figure}[t]
    \centering
    \includegraphics[width=\textwidth,
                 trim=0.5cm 0.6cm 0.6cm 0.7cm,
                 clip]{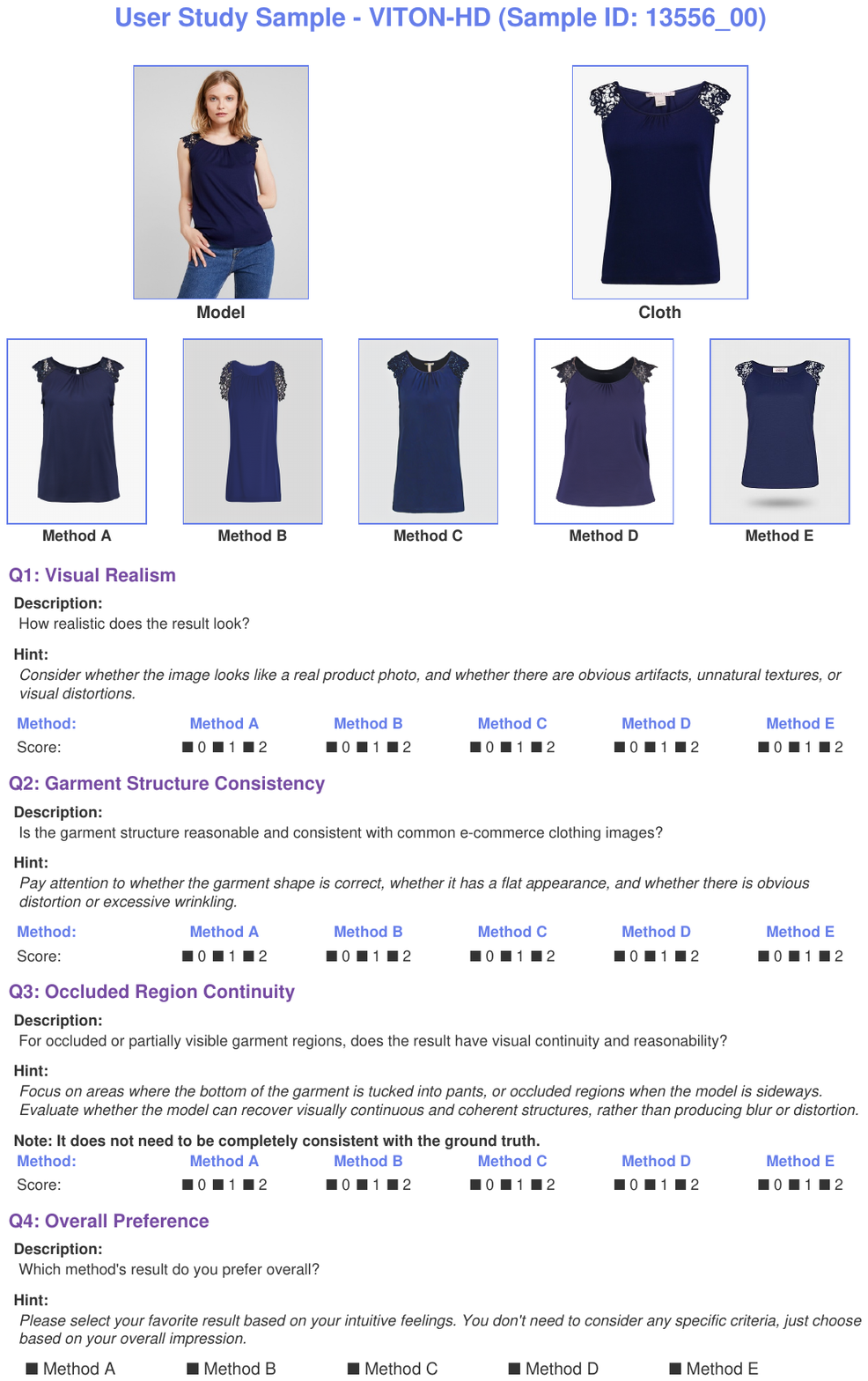}
    \caption{An example of the user study questionnaire for a single image set. 
    Participants are presented with multiple generated results for the same model image and are asked to rate each result based on visual realism, garment structure consistency, occluded region continuity, and overall preference.}
    \label{fig:user_study}
\end{figure}

To further evaluate the perceptual quality and garment structural consistency of our generated results, we conduct a user study using a questionnaire-based approach. 
The study is designed to be fully anonymous and adheres to common academic research ethics, collecting no personal or sensitive information. All participants are volunteers, and all results are recorded and analyzed anonymously. In the questionnaire, participants are presented with a series of images showing the output of different methods for the same input model image. 
The generated images are anonymized and shuffled to prevent bias, and multiple random seeds are used to ensure fair comparison across all state-of-the-art methods. Participants are asked to rate each result based on four criteria: 
\textbf{Visual Realism}, \textbf{Garment Structure Consistency}, \textbf{Occluded Region Continuity}, and \textbf{Overall Preference}. 
Ratings are provided on a simple three-point scale (0 -- Poor, 1 -- Acceptable, 2 -- Good), and overall preference is collected as a single-choice selection among the competing methods. 
Participants are instructed to evaluate the images subjectively, considering visual realism, plausibility of garment structure, and continuity in occluded regions. An example of the questionnaire interface for a single image set is shown in Figure~\ref{fig:user_study}, illustrating how participants are asked to rate each method's output.

\section{Textual Description Generation}
\label{sec:text_gen}
To obtain fine-grained and controllable garment descriptions from model images, we adopt a multi-stage text description generation pipeline assisted by multiple Multimodal Large Language Models (MLLMs).\begin{figure}[t]
    \centering
    \includegraphics[width=\textwidth,
                 trim=0.5cm 0.6cm 0.6cm 0.7cm,
                 clip]{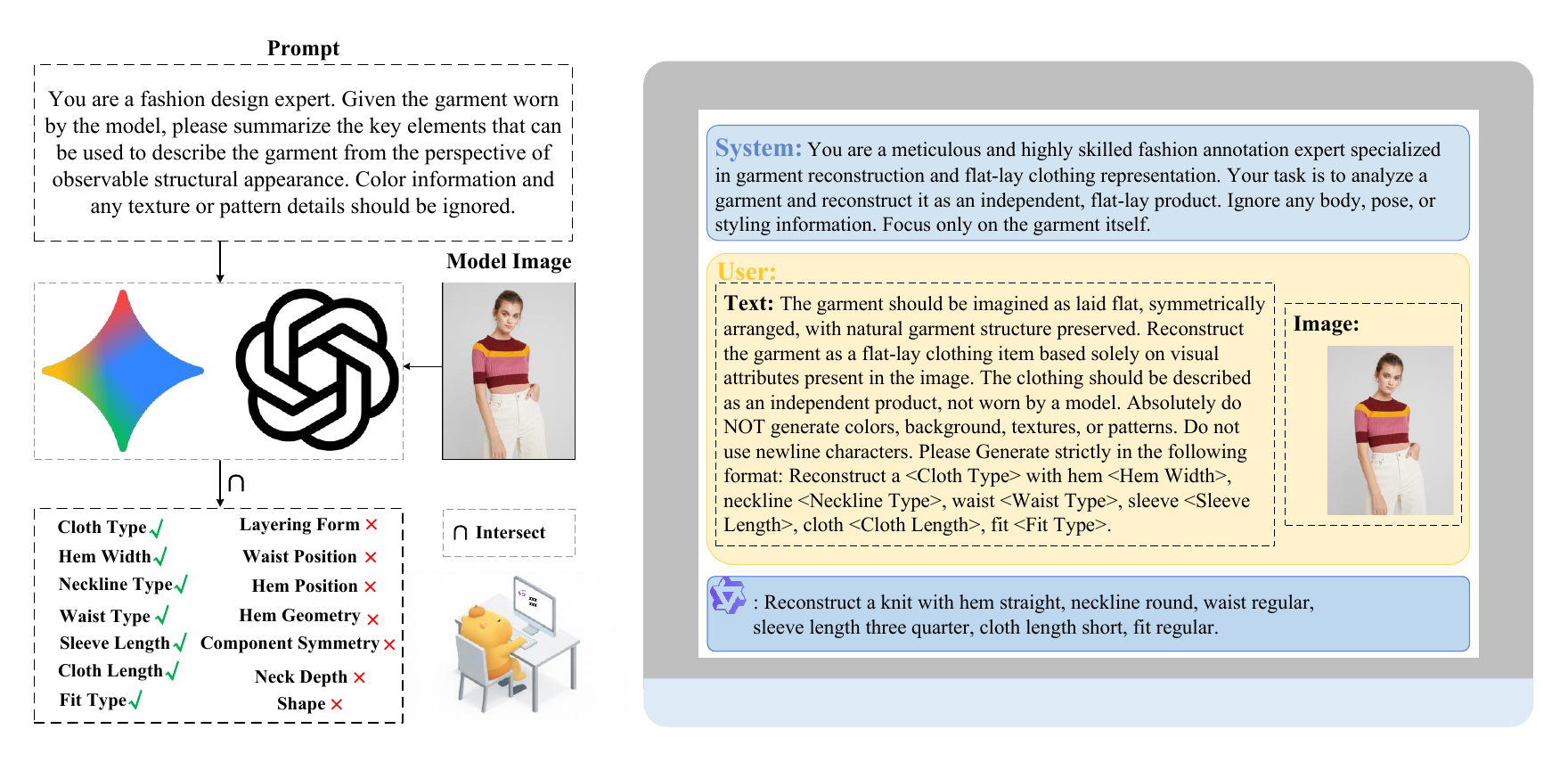}
    \caption{Overview of the textual description generation pipeline.
  \textbf{Left:} multiple MLLMs are employed to discover and filter visually grounded garment attributes.
  \textbf{Right:} the finalized structured attributes are used to guide Qwen3-VL for controllable garment description generation.}
  \label{fig:text_pipeline}
\end{figure}
Unlike approaches that rely on free-form natural language descriptions, we first aim to construct a visually grounded and semantically consistent structured garment attribute space. The attribute discovery and selection pipeline is illustrated in the \textbf{left} part of Fig.~\ref{fig:text_pipeline}.
Specifically, we introduce several state-of-the-art MLLMs, including ChatGPT     \cite{achiam2023gpt} and Gemini \cite{team2025gemma}, to collaboratively discover candidate garment attributes.
Given a model image, the models are prompted to analyze garment appearance elements that are visually discriminative and structurally observable, while explicitly excluding color, texture, and pattern information.
By aligning and comparing the candidate attributes proposed by different MLLMs, we retain only those attributes that reach consensus across multiple models and can be stably associated with visual appearance.
This process results in a compact yet expressive set of garment attributes, which serves as the foundation of the structured garment description space.
Rather than producing unconstrained text, Qwen3-VL is guided to generate attribute-consistent descriptions following the predefined template, ensuring interpretability and controllability of the generated results.
The interaction between the structured prompt and Qwen3-VL is shown in the \textbf{right} part of Fig.~\ref{fig:text_pipeline}
The final generated garment appearance description is denoted as $T_c$ \footnote{Here, we interpret $T_c$ as the garment appearance description prior to being processed by the text encoder. Although $T_c$ is defined in Table~\ref{tab:notation} as the appearance description of the garment after text encoding, we adopt this simplified interpretation to avoid introducing additional variables. The same notation convention is subsequently applied to $T_m$ and $T_{flat}$.}.

To accommodate different usage scenarios, we further perform lightweight text transformations on $T_c$.
Specifically, by replacing the leading keyword ``Reconstruct'' with ``A model is wearing'', we obtain a model-centric garment description $T_m$.
Subsequently, by removing the subject phrase, we derive a flat-lay oriented garment description $T_{flat}$, which follows a unified format of ``A flat-lay \textless category\textgreater'' and retains only the \textbf{Cloth Type} as the garment category indicator.
These transformations adapt the description to different semantic perspectives without altering the underlying attribute semantics.

\section{Qualitative Comparison with SOTA Methods}
\label{sec:add_qual}
\begin{figure}[h]
    \centering
    \includegraphics[width=0.8\textwidth]{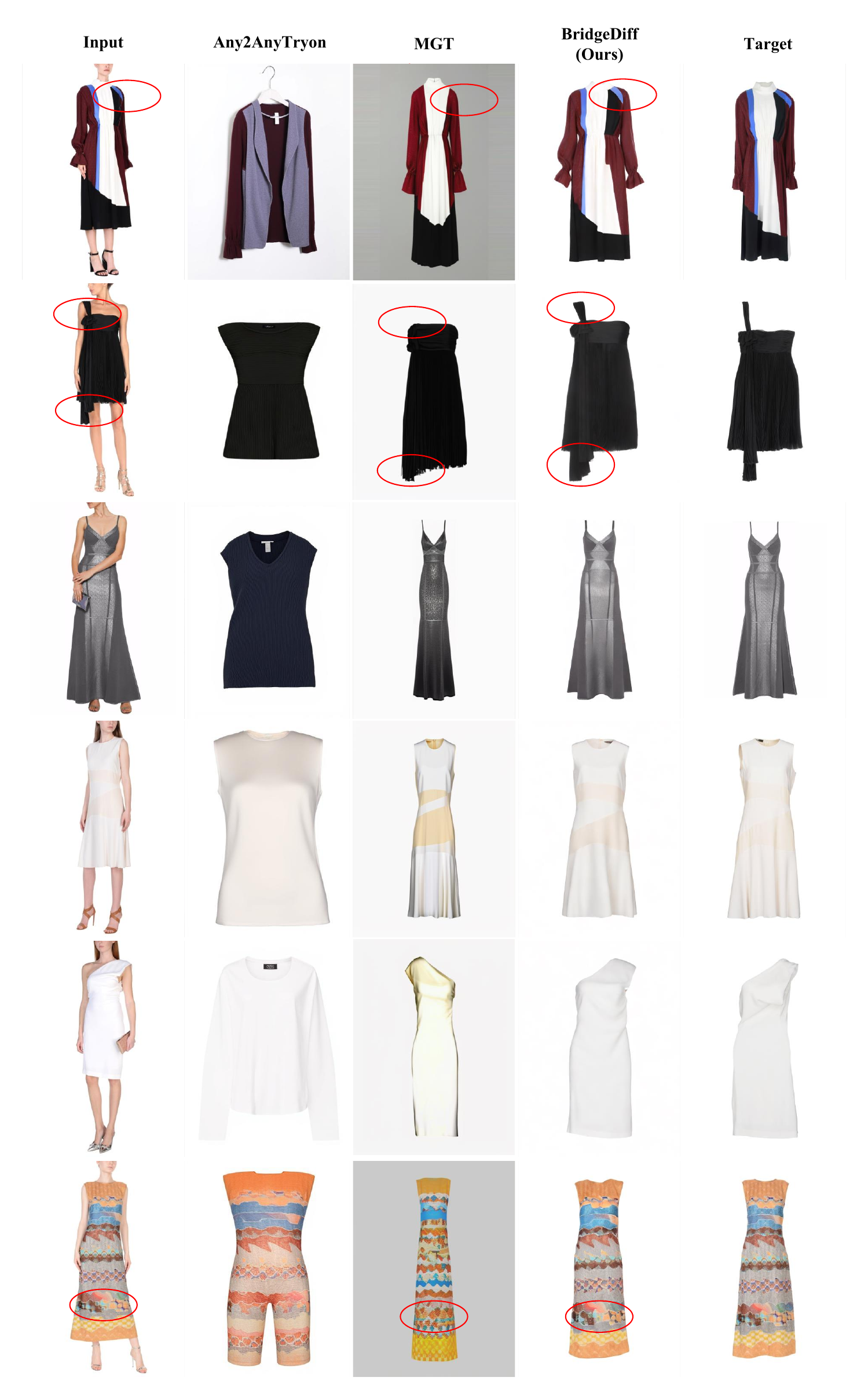}
    \caption{Additional qualitative comparisons on the DressCode dataset.
    \textcolor{red}{Red circles} highlight differences in local regions across different methods.  
    Unmarked examples indicate cases where the overall garment structure or color appearance differs from the reference.  
    \textbf{Zooming in provides a clearer view of these differences.}}
    \label{fig:dresscode_dressfin}
\end{figure}

\begin{figure}[h]
    \centering
    \includegraphics[width=0.8\textwidth]{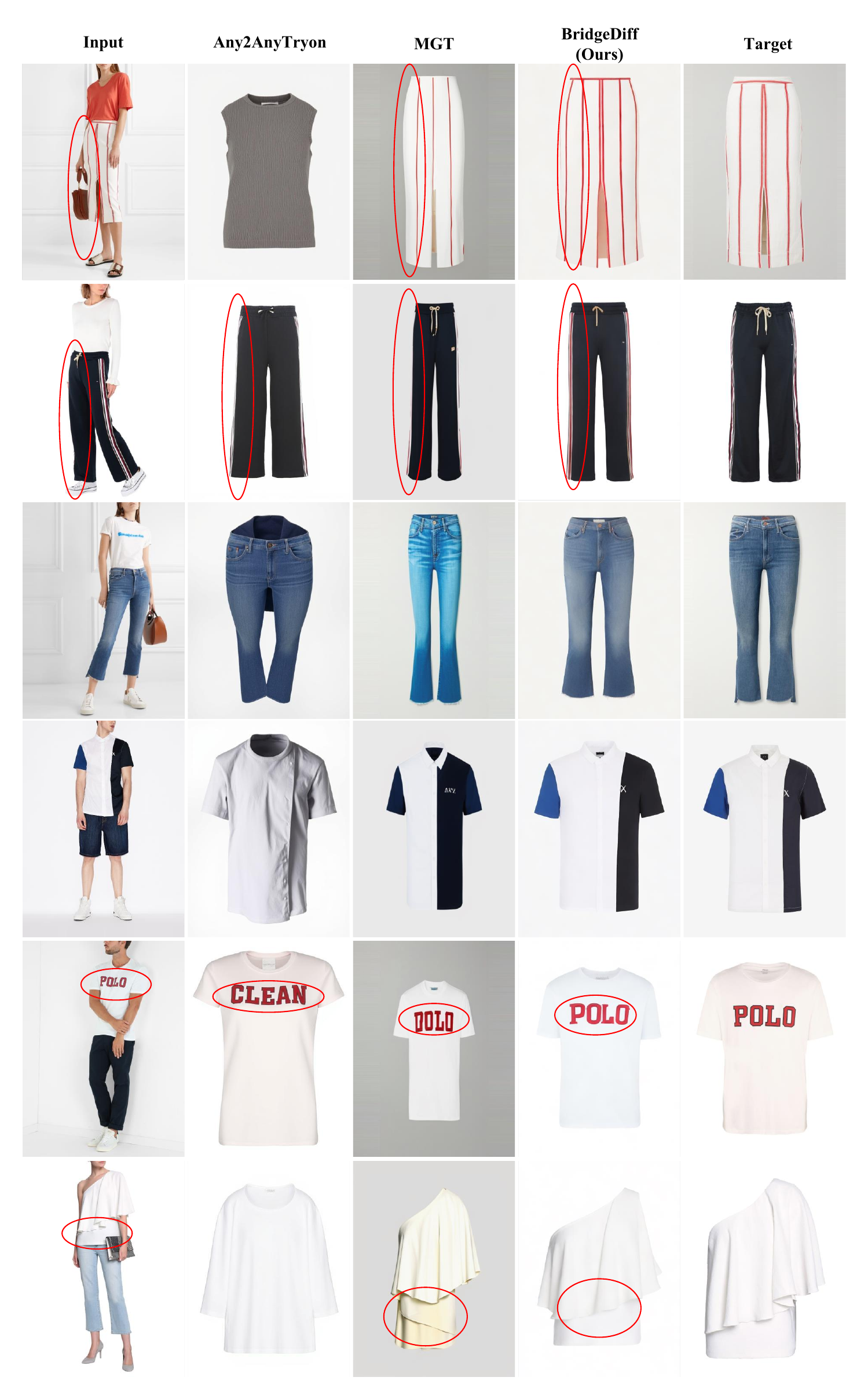}
    \caption{Additional qualitative comparisons on the DressCode dataset.
    \textcolor{red}{Red circles} highlight differences in local regions across different methods.  
    Unmarked examples indicate cases where the overall garment structure or color appearance differs from the reference.  
    \textbf{Zooming in provides a clearer view of these differences.}}
    \label{fig:dresscode_lianggefin}
\end{figure}

\begin{figure}[h]
    \centering
    \includegraphics[width=\textwidth,
                 trim=0.5cm 0.6cm 0.6cm 0.7cm,
                 clip]{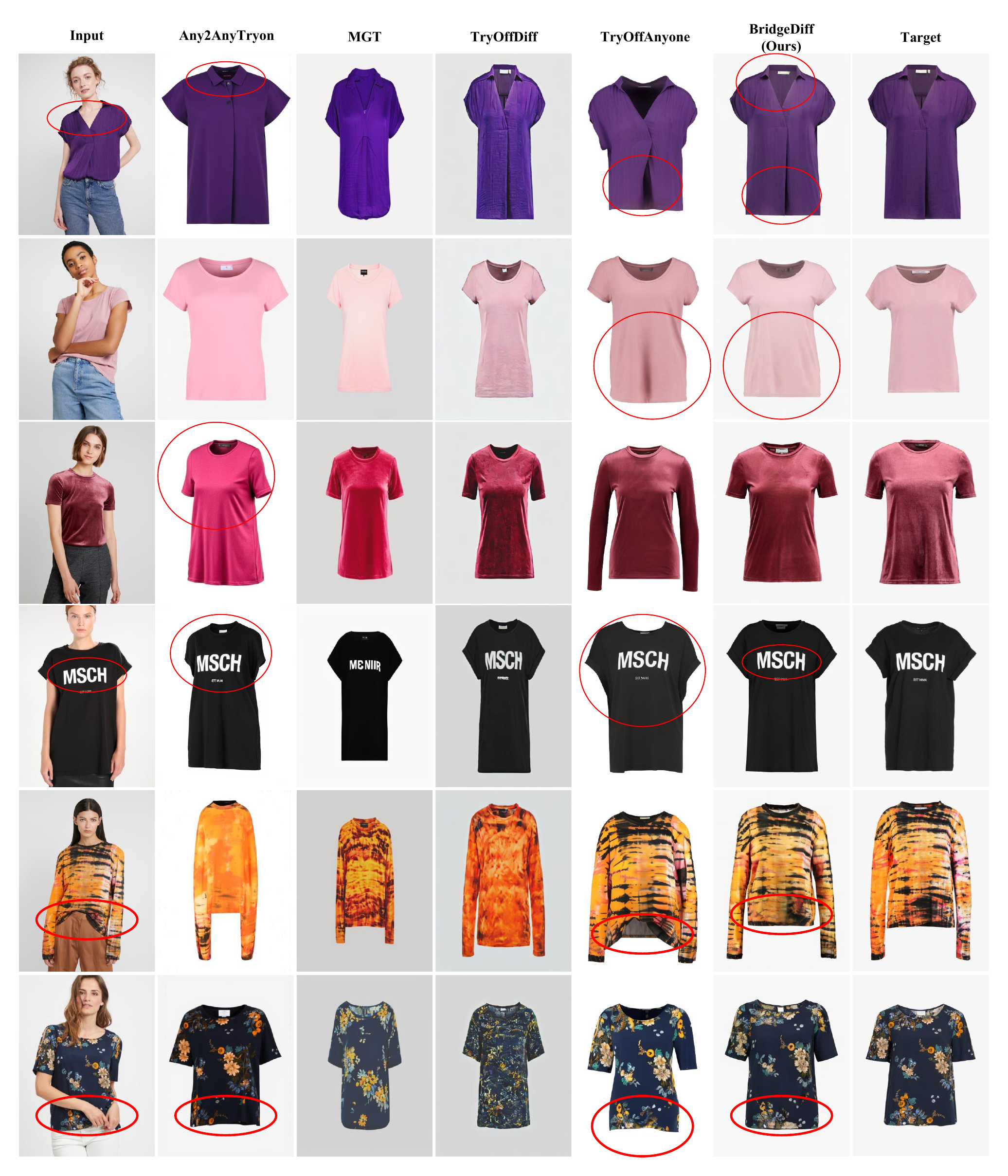}
    \caption{Additional qualitative comparisons on the VITON-HD dataset.
    \textcolor{red}{Red circles} highlight differences in local regions across different methods.  
    Unmarked examples indicate cases where the overall garment structure or color appearance differs from the reference.  
    \textbf{Zooming in provides a clearer view of these differences.}}
    \label{fig:supp_vitonhd_compared}
\end{figure}

To further evaluate our method, we provide additional qualitative comparisons with state-of-the-art (SOTA) methods on the DressCode and VITON-HD datasets.  
As shown in Fig.~\ref{fig:dresscode_dressfin}, we present results for Dresses on the DressCode dataset;  
Fig.~\ref{fig:dresscode_lianggefin} shows results for Lower-Body (first three rows) and Upper-Body (last three rows) on DressCode;  
and Fig.~\ref{fig:supp_vitonhd_compared} shows results on the VITON-HD dataset.

%
%

\end{document}